\documentclass[journal]{IEEEtran}

\usepackage{graphicx}  
\usepackage{array}        
\usepackage{multirow}     
\usepackage{booktabs}     
\usepackage{verbatim}     
\usepackage{subfiles}      
\usepackage{enumitem}
\setlist[enumerate,1]{label=\arabic*)}
\usepackage{makecell}
\usepackage{amsmath, amssymb}
\usepackage{soul}   
\usepackage[caption=false,font=normalsize,labelfont=sf,textfont=sf]{subfig}
\usepackage{algorithmic}
\usepackage[linesnumbered,ruled,vlined]{algorithm2e}

\SetNlSty{myglinenum}{}{:}
\usepackage{stfloats}     
\usepackage{url}        
\usepackage{textcomp}  
\usepackage[svgnames]{xcolor}
\usepackage[bookmarks=true]{hyperref}
\hypersetup{hidelinks,colorlinks=true,linkcolor=Blue,urlcolor=black,citecolor=Blue}
\newcommand{\etal}{et al.}
\newcommand{\eg}{e.g.}

\newcommand{\ie}{i.e.}

\begin{document}

\title{Skill-Aware Diffusion for Generalizable Robotic Manipulation}

\author{Aoshen Huang, Jiaming Chen, Jiyu Cheng,~\IEEEmembership{Member,~IEEE},
Ran Song,~\IEEEmembership{Senior Member,~IEEE},
Wei Pan,~\IEEEmembership{Member,~IEEE},
and~Wei Zhang,~\IEEEmembership{Senior Member,~IEEE}%
\thanks{This work was supported in part by the Shandong Provincial Natural Science Foundation under Grant ZR2025ZD12; in part by the National Natural Science Foundation of China under Grants 62573260 and U22A2057; and in part by the Key R\&D Program of Shandong Province under Grants 2025CXGC010111, 2024CXGC010212, and 2025CXGC010210. \textit{(Aoshen Huang and Jiaming Chen contributed equally to this work.) (Corresponding author: Jiyu Cheng.)}}%
\thanks{Aoshen Huang is with the School of Control Science and Engineering, Shandong University, Jinan 250061, China (e-mail: aoshenhuang@mail.sdu.edu.cn).}%
\thanks{Jiaming Chen is with the Department of Computer Science, The University of Manchester, Manchester M13 9PL, United Kingdom (e-mail: ppjmchen@gmail.com).}%
\thanks{Jiyu Cheng, Ran Song, and Wei Zhang are with the School of Control Science and Engineering, Shandong University, Jinan 250061, China, and are also with the Key Laboratory of Machine Intelligence and System Control, Ministry of Education, Jinan 250061, China (e-mail: jycheng@sdu.edu.cn; ransong@sdu.edu.cn; davidzhang@sdu.edu.cn).}%
\thanks{Wei Pan is with the School of Engineering, Newcastle University, Newcastle upon Tyne NE1 7RU, United Kingdom (e-mail: wei.pan2@newcastle.ac.uk).}%
}

\maketitle

        



\begin{abstract}
Robust generalization in robotic manipulation is crucial for robots to adapt flexibly to diverse environments. Existing methods usually improve generalization by scaling data and networks, but model tasks independently and overlook skill-level information. Observing that tasks within the same skill share similar motion patterns, we propose Skill-Aware Diffusion (SADiff), which explicitly incorporates skill-level information to improve generalization. SADiff learns skill-specific representations through a skill-aware encoding module with learnable skill tokens, and conditions a skill-constrained diffusion model to generate object-centric motion flow. A skill-retrieval transformation strategy further exploits skill-specific trajectory priors to refine the mapping from 2D motion flow to executable 3D actions. Furthermore, we introduce IsaacSkill, a high-fidelity dataset containing fundamental robotic skills for comprehensive evaluation and sim-to-real transfer. Experiments in simulation and real-world settings show that SADiff achieves good performance and generalization across various manipulation tasks. Dataset and videos are available at \url{https://sites.google.com/view/sa-diff}.
\end{abstract}

\renewcommand{\abstractname}{Note to Practitioners}
\begin{abstract}
This research aims to address the issue that existing methods tend to overlook shared skill-level motion patterns when generalizing robotic manipulation to diverse environments. The proposed approach introduces skill-level modeling to capture shared motion patterns in related tasks, improving adaptability to new scenarios. By combining skill-aware representations with a diffusion-based generation framework, the method produces more consistent motion flow even in unfamiliar scenarios. In addition, a skill-retrieval strategy further enhances the robustness of the mapping between motion flow and action. Our method has been validated through comprehensive experiments in simulated and real-world settings.
\end{abstract}


\section{Introduction}
\label{sec:intro}
\IEEEPARstart{L}{anguage}-conditioned robotic manipulation~\cite{zhao2024vlmpc,mees2022matters,li2025language}, which involves planning executable actions according to visual observations and language instructions, has gained increasing attention. As one of the primary paradigms in this field, imitation learning has demonstrated strong potential by enabling robots to acquire manipulation skills directly from collected demonstrations~\cite{ma2025contrastive,lindata}. However, achieving robust generalization across various objects and environments remains a persistent challenge~\cite{chen2025semantically}, as the classic imitation learning paradigm~\cite{zhang2018deep,zhao2023act} depends heavily on task-specific data and struggles to adapt to distributional shifts, thus affecting its adaptability and flexibility in unfamiliar scenarios.

To improve generalization in robotic manipulation, previous research has explored two main directions. The first direction enhances visual representations within end-to-end imitation learning frameworks through large-scale pretraining and auxiliary learning objectives~\cite{mavip,nair2023r3m,zeng2024learning}. For instance, Nair \etal~\cite{nair2023r3m} adopted time-contrastive learning to capture robust visual patterns, while Zeng \etal~\cite{zeng2024learning} leveraged interaction prediction to model behavioral dynamics. However, these methods generally require large-scale datasets, and the learned visual representations may retain task-irrelevant elements (\eg, background details or textures), which can reduce the efficiency and accuracy of downstream action prediction. The second direction formulates imitation learning as a two-stage framework, which first learns to predict future task-relevant motion representations (\eg, task videos~\cite{bharadhwaj2024gen2act}, goal images~\cite{ni2024generate}, or scene flow~\cite{track2act}) from easily accessible video data, and then leverages these representations as auxiliary inputs for visuomotor policy learning or directly maps them to executable actions. Among various motion representations, flow-based methods~\cite{im2flow2act,chen2025ec} are particularly effective at capturing task-relevant motion dynamics while suppressing irrelevant visual distractions. However, attaining strong generalization with these methods remains dependent on large-scale training data. Moreover, such approaches face challenges in achieving a balance between robustness and precision when translating predicted 2D motion representations into executable 3D actions. For instance, learning-based strategies~\cite{track2act,im2flow2act} are sensitive to intrinsic and extrinsic camera parameters, while vanilla heuristic-based strategies~\cite{chen2025ec,ko2023actionless} suffer from unstable performance due to sensor noise. To address these limitations, we exploit shared skill-level knowledge across diverse tasks to achieve data-efficient generalization without relying on large-scale data.

Inspired by the fact that different tasks within the same skill domain exhibit similar motion patterns~\cite{fang2021discovering,liang2024skilldiffuser,yao2025think}, we exploit skill-level information as additional guidance and propose the Skill-Aware Diffusion (SADiff) framework to improve generalization in robotic manipulation. First, a skill-aware encoding module uses learnable skill tokens to capture skill-specific information while interacting with multimodal inputs. Then, a skill-constrained diffusion model generates object-centric motion flow conditioned on the produced skill-aware token sequences, allowing the model to capture unified features across diverse tasks within the same skill domain while preserving task-specific information. We further introduce two auxiliary objectives, namely the skill classification loss and the skill contrastive loss, to distinguish the motion patterns associated with different skills and improve the accuracy and robustness of motion flow generation. Finally, we propose a skill-retrieval transformation strategy that translates the generated 2D motion flow to executable 3D actions by retrieving skill-specific priors to refine the 2D-to-3D mapping. By systematically incorporating skill-level information throughout the framework, SADiff achieves strong generalization across diverse tasks and environments.

Furthermore, existing robotic manipulation datasets~\cite{yu2020meta,james2020rlbench,liu2023libero} are primarily designed to evaluate task-level completion and therefore lack the granularity required to assess specific manipulation skills. To address this limitation, we construct IsaacSkill, a new dataset specifically designed for skill-centric evaluation. Built on the high-fidelity NVIDIA Isaac Lab platform~\cite{isaaclab}, IsaacSkill encompasses a set of fundamental robotic skills instantiated across a diverse range of manipulation tasks. The dataset enables a comprehensive evaluation of generalization while providing physically realistic dynamics that support robust sim-to-real transfer.

In summary, the contributions of this work are as follows:
\begin{enumerate}[label=\arabic*)]
    \item We propose the SADiff framework to improve generalization in robotic manipulation by explicitly modeling skill-level representations and systematically integrating them into the encoding, generation, and execution phases.
    \item We develop a skill-constrained diffusion model that generates object-centric motion flow conditioned on skill-aware token sequences, with skill contrastive learning further improving the robustness of flow generation.
    \item We design a skill-retrieval transformation strategy that uses skill-specific priors to refine the mapping from predicted object-centric 2D motion flow to executable 3D actions, improving precision and consistency without additional training.
    \item We construct IsaacSkill, a new dataset on the high-fidelity NVIDIA Isaac Lab platform, covering fundamental skills across diverse tasks to enable skill-centric evaluation and facilitate zero-shot sim-to-real transfer.
\end{enumerate}

\section{Related Work}
\label{sec:relatedwork}

\subsection{Learning from Demonstrations in Robotic Manipulation}
Imitation learning, which enables robots to directly learn from collected demonstrations, has recently become a primary approach in robotic manipulation~\cite{zhao2023act,kroemer2021review,drolet2024comparison,li2025robotic,mandlekar2021matters,chi2023diffusion}. In the standard imitation learning framework, a policy network is trained to replicate expert behavior by learning a mapping from state observations to the corresponding actions~\cite{mandlekar2021matters}. For instance, ACT~\cite{zhao2023act} employed a conditional variational autoencoder with a Transformer architecture to predict multi-step action chunks from visual observations, enabling effective manipulation with a few demonstrations. Chi \etal~\cite{chi2023diffusion} proposed diffusion policy to model complex action distributions through an iterative denoising diffusion process, efficiently handling the multimodality of trajectories in collected demonstrations.

To improve the ability of robots to understand and follow human instructions, language-conditioned imitation learning has been increasingly explored for robotic manipulation, where visual inputs and language instructions are integrated to plan executable actions~\cite{zhou2023language,stepputtis2020language,yao2025long}. For instance, CLIPort~\cite{shridhar2022cliport} leveraged a pretrained CLIP~\cite{clip} model to align visual and language features, allowing robots to directly interpret and perform language-instructed tasks. PerAct~\cite{shridhar2023perceiver} employed a perceiver transformer~\cite{jaegle2021perceiver} to integrate language features with voxelized RGB-D observations for precise action prediction. Zhang \etal~\cite{zhang2026language} proposed a mixture-of-experts diffusion policy that incorporates semantically grounded visual representations to mitigate perceptual ambiguity in multi-task robotic manipulation. However, these methods typically suffer from poor generalization in unseen scenarios due to their heavy reliance on extensive, task-specific demonstrations. Crucially, most existing approaches learn individual tasks independently and do not explicitly exploit shared motion patterns across related tasks, which limits their ability to transfer previously acquired knowledge to new tasks and environments.

\subsection{Improving Generalization in Imitation Learning}
To improve generalization in imitation learning for robotic manipulation, one intuitive direction is to enhance visual encoders by pretraining on large-scale manipulation-related datasets~\cite{grauman2022ego4d,o2024open}  within an end-to-end imitation learning framework~\cite{mavip,nair2023r3m,zeng2024learning,Xiao2022}. For instance, R3M~\cite{nair2023r3m} pre-trained a visual encoder on human video data, using time-contrastive learning and video-language alignment to obtain robust visual features. MVP~\cite{Xiao2022} employed masked image reconstruction to learn visual representations for generalized robotic motion control. MPI~\cite{zeng2024learning} introduced interaction prediction as an auxiliary objective to capture behavioral patterns and physical interactions, improving the generalization of the visuomotor policy. MAPLE~\cite{gavryushin2025maple} learned manipulation-specific visual priors by predicting future hand-object contact points and 3D hand poses. However, these methods typically rely on large-scale datasets for pretraining, which requires labor-intensive demonstration collection and massive computational resources. Moreover, the learned visual representations may retain task-irrelevant information, which can negatively affect downstream action prediction.

Another major direction formulates imitation learning as a two-stage framework~\cite{bharadhwaj2024gen2act,track2act,wen2023any}. In the first stage, the model predicts a future task-relevant motion representation, such as task videos \cite{bharadhwaj2024gen2act}, goal images~\cite{ni2024generate}, or scene flow \cite{track2act}. In the second stage, the predicted representation is either used as an auxiliary condition for visuomotor policy learning or directly converted into executable actions. Among diverse motion representations, motion flow in the 2D pixel space has become a powerful intermediate representation in motion planning. It can be easily extracted from raw videos using advanced point tracking methods and learned from large-scale video datasets. For instance, Track2Act~\cite{track2act} employed a goal-conditioned grid flow prediction model to generate future trajectories of query points and converted these trajectories into residual actions via a policy network. Im2Flow2Act~\cite{im2flow2act} leveraged object-centric motion flow to capture the dynamics of the target objects and translated the motion flow into actions via a flow-conditioned policy network, facilitating embodiment-agnostic manipulation. General Flow~\cite{yuan2024general} formulated language-conditioned 3D flow derived from RGB-D human videos as embodiment-agnostic affordances to guide closed-loop robot control. However, existing approaches typically treat tasks in isolation, thereby overlooking motion priors shared across related tasks and neglecting to explicitly leverage skill-level information.

\subsection{Diffusion Models in Robotic Manipulation}
Diffusion models have recently demonstrated considerable potential in robotic manipulation~\cite{wolf2025diffusion, zhang2025generative}. Through iterative denoising, diffusion-based approaches can effectively capture the multimodal distributions of expert demonstrations and generate smooth and coherent behaviors~\cite{pearce2023imitating}. Previous works introduced denoising diffusion probabilistic models into visuomotor control by modeling the conditional distribution of robot actions given visual observations, achieving strong performance across various tasks~\cite{chi2023diffusion,reuss2023goal}. Subsequent works extended this paradigm by incorporating richer sensory inputs (\eg, 3D visual representations~\cite{Ze2024DP3} and tactile feedback~\cite{xue2025reactive}), as well as more expressive geometric and semantic conditioning mechanisms~\cite{chen2025g3flow}. These advances have further enhanced the multimodal perception and spatial understanding capabilities of diffusion-based visuomotor policies.

Beyond directly generating robot actions, diffusion models can also synthesize future videos as visual plans for robotic manipulation~\cite{ko2023actionless,du2023learning}. For instance, UniPi~\cite{du2023learning} employed a language-conditioned video diffusion model to predict future frames and subsequently recovered executable actions using an inverse dynamics model. LVP~\cite{chen2025large} generated zero-shot video plans conditioned on an initial observation and task instruction, and extracted actions through 3D hand reconstruction and embodiment-specific retargeting. However, video-based methods must jointly model task-relevant motion and appearance details, increasing generation complexity and making the resulting plans susceptible to irrelevant visual variations. In contrast, flow-based methods explicitly capture object-centric motion, providing a more compact and direct representation of task dynamics.

\subsection{Imitation Learning with Skill Representations}
Existing studies have explored various ways of incorporating skill representations into policy learning frameworks~\cite{liang2024skilldiffuser,shi2025zeromimic,Zhao11580387}. One branch of research treats each skill as a task or subtask, using it as a high-level specification of what the policy should accomplish. Such representations partially or entirely replace the original task inputs, abstracting task semantics and facilitating task decomposition, skill reuse, and compositional execution. For example, Xskill~\cite{xu2023xskill} employed self-supervised clustering to encode multi-task video demonstrations into sequences of skill embeddings, where each embedding represents a semantically meaningful subtask that can be reordered across different procedures. SkillDiffuser~\cite{liang2024skilldiffuser} introduced vector quantization (VQ) to discretize task inputs into a set of latent skills, thereby decomposing complex manipulation objectives into reusable subgoals and facilitating hierarchical task execution. Another branch represents skills as basic motion behaviors or atomic actions, such as ``\textit{Move up}'' or ``\textit{Rotate the gripper}''. Rather than specifying the task objective, these skills characterize how the robot should move and are incorporated into the policy as reusable low-level action primitives. For instance, UniSkill~\cite{kim2025uniskill} learned motion-centric skill representations from unlabeled videos through inverse and forward skill dynamics models and incorporated them into policy learning. Fang \etal~\cite{fang2025learning} represented actions using optical flow and applied finite scalar quantization (FSQ) to extract discrete skill codes corresponding to distinct motion patterns from video demonstrations. In addition, Gu \etal~\cite{gu2026learning} proposed a skill-conditioned diffusion policy in which a lightweight routing network dynamically selects and assigns appropriate motion primitives to individual states.

Despite their effectiveness, existing formulations generally overlook the behavioral structures shared by tasks within the same skill category. For example, tasks associated with the same semantic skill (\eg, ``\textit{Picking \& Placing}'') often exhibit similar motion patterns despite variations in objects and environments. In contrast to prior methods, we explicitly represent such skill-level information using a set of learnable skill tokens. These tokens are designed to capture motion patterns shared across tasks within each skill category and are combined with task-specific inputs to guide policy learning, thereby improving generalization in unseen scenarios.

\begin{figure*}[t]
    \centering
    \includegraphics[width=1.0\textwidth]{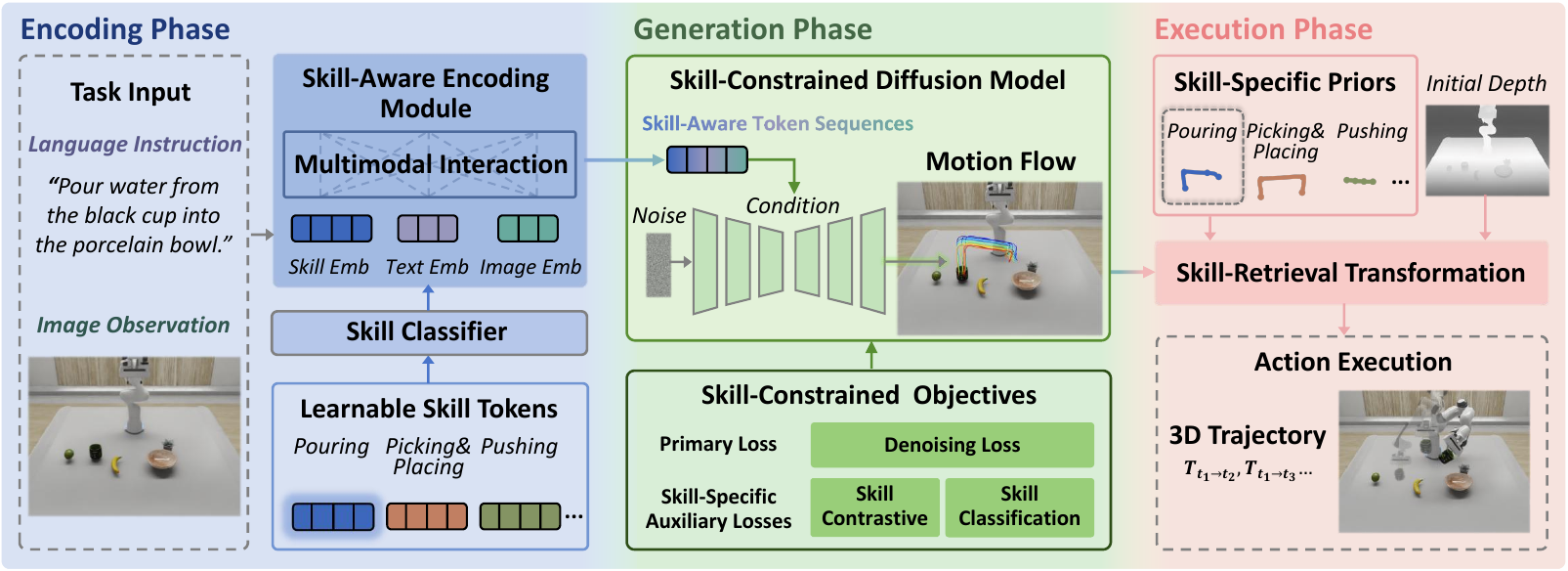}
    \caption{Overview of the proposed Skill-Aware Diffusion (SADiff) framework. It is structured into three phases: (1) The encoding phase, where the skill-aware encoding module uses learnable skill tokens to interact with multimodal inputs and extract skill-specific information; (2) The generation phase, in which a skill-constrained diffusion model generates object-centric motion flow conditioned on the skill-aware token sequences and is optimized using a denoising loss and two skill-specific auxiliary losses; and (3) The execution phase, which employs a skill-retrieval transformation strategy to translate the generated 2D motion flow into executable 3D trajectories by leveraging skill-specific priors.}
    \label{fig_method}
\end{figure*}

\section{Preliminaries}
\label{sec:preliminary}
\subsection{Problem Formulation}
Following existing works such as Im2Flow2Act~\cite{im2flow2act} and Track2Act~\cite{track2act}, we decompose the robotic manipulation problem into two sub-problems. The first is flow generation, which aims to interpret the task from an image observation and a language instruction, and produce 2D motion flow that describes the pixel-level movement of the target object over a time horizon. The second sub-problem is to lift the 2D motion flow into a 3D trajectory and convert it into a sequence of executable actions through a flow transformation stage.

Formally, given an initial image observation $I$ and a language instruction $L$, the first sub-problem is to learn a flow generator $\mathcal{G}$ that maps them to a 2D motion flow $F$:
\begin{equation}
    F=\mathcal{G}(I,L)
\end{equation}
where $F = \{ P_1, P_2, \dots, P_T \}$ represents the pixel-wise motion of the target object across $T$ time steps and $P_t$ denotes a set of 2D coordinates of keypoints on the target object at time step $t$. The second sub-problem is to design a transformation strategy $\mathcal{T}$ that maps the predicted motion flow $F$ to an executable action sequence $A$:
\begin{equation}
A = \mathcal{T}(F)
\end{equation}
where $A = \{ a_1, a_2, \dots, a_T \}$ denotes the action sequence of the end-effector over the task horizon, and each action $a_t$ represents the 6-DoF end-effector pose at time step $t$. We detail the proposed approach in the following section.

\subsection{Motion Flow Extraction}
\label{extracte_flow}
We adopt a motion flow extraction paradigm similar to that of Im2Flow2Act~\cite{im2flow2act}.
First, we employ Qwen-VL~\cite{Qwen-VL} as an open-set detector to identify and locate the target object in the first frame of the video demonstration and obtain the corresponding bounding box. Then we uniformly sample a set of keypoints $P_1 \in \mathbb{R}^{2 \times N_x \times N_y}$ within the bounding box, where the first channel represents the 2D coordinates of the points, and $N_x \times N_y$ indicates the number of sampled points. Next, we employ TAPIR~\cite{doersch2023tapir} as a keypoint tracker, which tracks the motion of each keypoint across the entire video demonstration, yielding the motion flow of the target object. The extracted motion flow can be denoted as $F = \{ P_1, P_2, \dots, P_T \} \in \mathbb{R}^{2 \times N_x \times N_y \times T}$, where each $P_t$ corresponds to the tracked keypoints in frame $t$, and $T$ is the total number of frames in the video. The extracted motion flow serves as the training sample for the diffusion model.

\section{Method}
\label{sec:method}
Fig.~\ref{fig_method} presents an overview of SADiff. Its key components are detailed as follows: Section~\ref{sec:encoding} introduces the skill-aware encoding module, which employs a set of learnable skill-aware tokens to capture skill-specific information while interacting with multimodal task inputs. Section~\ref{sec:flow_generation} introduces the skill-constrained diffusion model that generates the motion flow of the target object conditioned on the skill-aware token sequences. We adopt a multi-objective training paradigm for flow generation and introduce a skill contrastive objective to improve robustness. Finally, Section~\ref{sec:retrieval} details the skill-retrieval transformation strategy, which uses skill-specific trajectory priors to guide the optimization-based transformation process, accurately converting the 2D motion flow into executable 3D actions.

\subsection{Skill-Aware Encoding}
\label{sec:encoding}
\begin{figure}[t]
    \centering
    \includegraphics[width=1\linewidth]{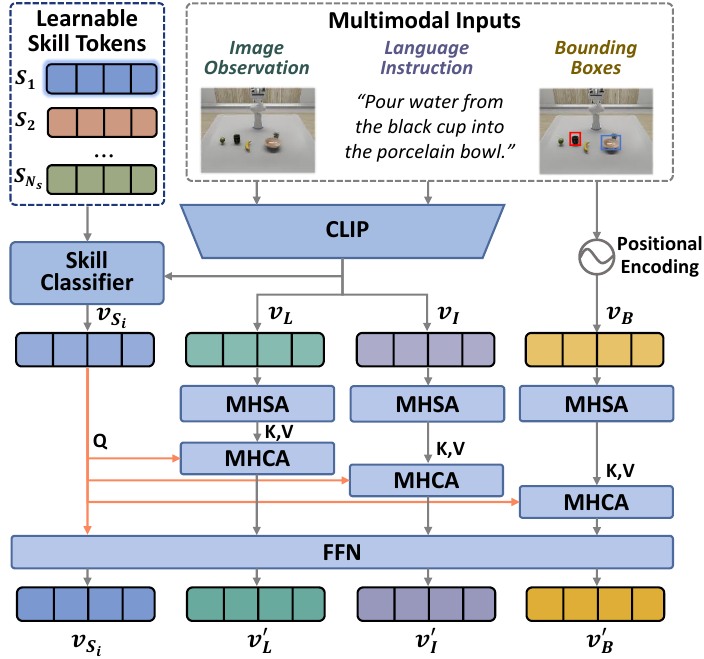}
    \caption{Architecture of the skill-aware encoding module. The skill-aware encoding module integrates image, language, and bounding boxes of relevant objects with learnable skill tokens through attention-based interactions, producing skill-aware token sequences.}
    \label{fig_SAE}
\end{figure}

We design a skill-aware encoding module that facilitates interactions among multimodal inputs and extracts skill-specific information, as illustrated in Fig.~\ref{fig_SAE}. The encoding process begins with an RGB image captured by a camera, represented as $I\in\mathbb{R}^{H\times W\times 3}$. Given a language instruction $L$, such as \textit{``Pour water from the black cup into the porcelain bowl''}, we use an advanced vision-language model, \ie, Qwen-VL~\cite{Qwen-VL}, to identify and locate the target objects \textit{``black cup''} and \textit{``porcelain bowl''}, extracting their bounding boxes $B \in\mathbb{R}^{N_o\times 4}$, where $N_o$ represents the number of related objects. To capture skill-specific information and facilitate dynamic interactions across multimodal inputs, we introduce a set of learnable skill tokens, which are randomly initialized and represented as $S \in\mathbb{R}^{N_s\times D}$, where $N_s$ denotes the number of skills. Each row of $S$ corresponds to one skill category, and the complete embedding matrix is registered as a trainable parameter and jointly optimized with the subsequent diffusion model. For a given task, the skill-relevant token $S_{\hat{i}} \in \mathbb{R}^{D}$ is selected from $S$ using an MLP-based skill classifier conditioned on the image observation $I$ and language instruction $L$, where $\hat{i}$ denotes the predicted skill index. Furthermore, we encode $I$, $L$, $B$, and $S_i$ independently to obtain initial token sequences $v_I$, $v_L$, $v_B$, $v_{S_i}$, all of which are transformed to a common encoding dimension $D$. Specifically, we use the pretrained CLIP~\cite{clip} to encode both the image and language instruction. The bounding boxes are encoded using 2D sinusoidal positional encoding~\cite{vaswani2017attention}, capturing spatial relationships. The skill tokens are encoded using a simple fully connected layer.

To generate semantically rich skill representations and enable effective information exchange across multimodal token sequences, we employ multi-head self-attention (MHSA) and multi-head cross-attention (MHCA). These mechanisms enhance the interaction of image, text, and bounding box tokens with skill tokens, facilitating the alignment and fusion of multimodal information. For conciseness, the MHSA and MHCA mechanisms are formally defined as follows:
\begin{equation} 
\text{MHSA}(X) = \sigma(f_{Q}(X)f_{K}(X)^\mathsf{T})f_{V}(X) 
\end{equation}
\begin{equation} 
\label{MHCA} 
\text{MHCA}(X, Y) = \sigma(f_{Q}(X)f_{K}(Y)^\mathsf{T})f_{V}(Y) 
\end{equation}
where $\sigma$ is the softmax function, while $f_Q$, $f_K$, and $f_V$ represent the query, key, and value mapping functions, respectively.

The multimodal token sequences $v_I, v_L,v_B$ first undergo initial self-encoding via MHSA, after which the skill token sequence acts as the query input, while the remaining token sequences serve as the key and value inputs in MHCA. This process is followed by a feed-forward network (FFN) implemented with multi-layer perceptrons. The interaction process is expressed as:
\begin{equation}
\label{encode}
v'_{x} = \text{FFN}(\text{MHCA}(v_{S_i}, \text{MHSA}(v_x))) 
\end{equation}
where $x \in \{I, L, B\}$. The resulting skill-aware token sequences $v_{S_i}, v'_{I},v'_{L},v'_{B}$ are used as conditioning inputs for the diffusion model in the subsequent flow generation stage. Unlike existing methods~\cite{im2flow2act} that often process multimodal inputs independently and rely on static feature alignment, our skill-aware encoding module introduces learnable skill tokens and leverages MHSA and MHCA mechanisms to dynamically align and fuse skill-specific features across modalities. Through this encoding process, the resulting skill-aware token sequences capture shared features across diverse tasks within the same skill category, while preserving task-specific details, thereby providing an effective condition for the following skill-constrained diffusion model.

\subsection{Skill-Constrained Flow Generation}
\label{sec:flow_generation}
To generate a precise 2D object-centric motion flow that is aligned with a specific skill, the current task, and the target objects, we propose a novel skill-constrained diffusion model. This model is equipped with a denoising loss to supervise flow generation and two auxiliary skill-specific losses that guide the model to capture semantic cues related to the skill. We first introduce the diffusion model for flow generation, and then explain how we design and leverage the denoising loss and the two auxiliary losses to train the model.

\subsubsection{Diffusion Model for Flow Generation}

First, the motion flow $F$ is encoded into a latent representation space using a pretrained VAE~\cite{kingma2013auto} encoder $E_\phi(\cdot)$, denoted as:
\begin{equation}
    z_{0}=E_\phi(F)
\end{equation}
where $z_{0}$ is the initial latent representation of the motion flow. Then the latent representation $z_{0}$
is perturbed over $k$ steps using the forward diffusion process to derive the perturbed latent representation $z_{k}$:
\begin{equation}
z_{k}=\sqrt{\overline\alpha_k}z_{0} +\sqrt{1-\overline{\alpha}_k}\epsilon, \epsilon\sim \mathcal{N}(0,I)\
\end{equation}
where $\bar{\alpha}_k$ is the cumulative noise schedule at step $k$ and $\mathcal{N}(0,I)$ is a Gaussian noise distribution.

Subsequently, $z_{k}$ is processed by a UNet-based~\cite{unet}  noise prediction network $\epsilon_\theta$ to predict the noise added during the forward diffusion process, conditioned on skill-aware token sequences $\textbf{v}_{S_i}=[v_{S_i}, v'_{I},v'_{L},v'_{B}]$:
\begin{equation}
\hat\epsilon=\epsilon_{\theta}(z_{k},\textbf{v}_{S_i})
\end{equation}
where $[\cdot]$ denotes the concatenation operation, and $\hat{\epsilon}$ represents the predicted noise. Following AnimateDiff~\cite{guo2023animatediff} and Im2Flow2Act~\cite{im2flow2act}, we integrate an MHSA layer as a motion module into $\epsilon_\theta$ that performs information exchange on each spatial feature along the temporal dimension to capture temporal dependencies across frames. 

During inference, the reverse diffusion process starts by sampling random noise $\hat{z}_{m} \sim \mathcal{N}(0, I)$ in the latent space. We then perform $m$ denoising steps to generate the latent $\hat{z}_0$. Finally, a pretrained decoder $D_\phi(\cdot)$ maps the latent $\hat{z}_0$ back to pixel space to generate object-centric motion flow $\hat{F}$:
\begin{equation}
\hat{F} = D_\phi(\hat{z}_0).
\end{equation}

\subsubsection{Skill-Constrained Training}

\begin{figure}[t]
    \centering
    \includegraphics[width=1\linewidth]{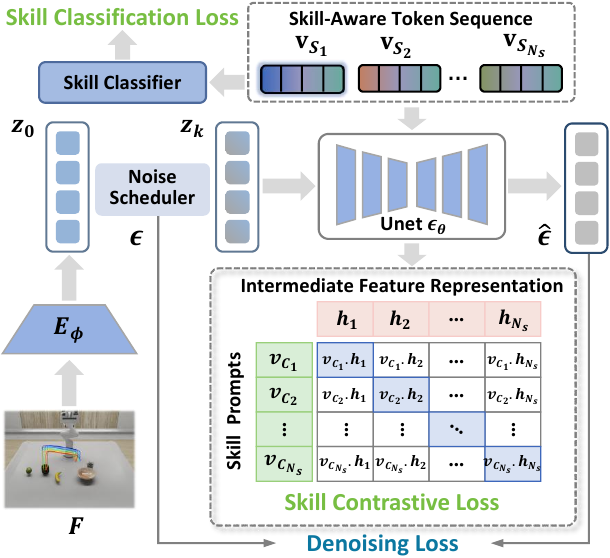}
    \caption{Overview of the skill-constrained diffusion model. The diffusion model generates motion flow by jointly optimizing skill classification loss, skill contrastive loss, and denoising loss to ensure accurate skill selection, semantic alignment, and precise flow generation.}
    \label{fig_con}
\end{figure}

To ensure that the generated motion flow is precise and aligned with the intended skill, we train the diffusion model with a combination of the primary denoising loss for flow generation and two skill-specific auxiliary losses. Fig.~\ref{fig_con} provides an overview of the skill-constrained diffusion model and its optimization objectives.

\paragraph*{Denoising Loss}  
Consistent with previous diffusion-based approaches~\cite{im2flow2act,guo2023animatediff}, we adopt a mean squared error (MSE) loss between the predicted noise $\hat{\epsilon}$ and the ground-truth noise $\epsilon$:
\begin{equation}
    \mathcal{L}_{\text{MSE}} = \left\| \hat{\epsilon} - \epsilon \right\|^2.
\end{equation}
This objective facilitates precise denoising by minimizing the divergence between predicted and ground-truth noise.

\paragraph*{Skill Classification Loss}  
To enable the skill-aware encoding module to select the optimal skill-relevant token from the full set of learnable skill tokens $S \in\mathbb{R}^{N_s\times D}$, we design an auxiliary classification branch based on multi-layer perceptrons. Specifically, the image feature $v_I$ and the language feature $v_L$ encoded by the pretrained CLIP encoder are concatenated and passed through the MLP-based skill classifier to predict the skill category $\hat{i}$, which indicates the most relevant skill token $S_{\hat{i}} \in \mathbb{R}^{D}$. During both training and inference, token retrieval is based on the predicted skill index rather than the ground-truth skill label. The ground-truth label is used only to supervise the skill classifier through the cross-entropy loss. The selected token $S_{\hat{i}}$ is then used in the subsequent multimodal encoding stage as described in Section~\ref{sec:encoding}, ensuring that the skill-aware token sequences are produced correctly. The classification loss is defined as:
\begin{equation}
\mathcal{L}_{\text{CE}} = - \sum_{i=1}^{N_s} y_i \log(p_i)
\end{equation}
where $y_i$ represents the one-hot encoded ground-truth skill label and $p_i$ denotes the predicted probability of the $i$-th skill category.

\paragraph*{Skill Contrastive Loss}
To further ensure that the diffusion model distinguishes between motion patterns of different skills and aligns them with their semantic descriptions, we propose a skill contrastive loss that aligns the intermediate features of $\epsilon_\theta$ with the semantic representations of the pre-defined skill prompts. Specifically, we construct a skill prompt bank $v_C \in \mathbb{R}^{N_s \times D}$ by encoding high-level skill categories (e.g., \textit{``Pouring''}, \textit{``Picking and Placing''}) using the CLIP text encoder. Both the CLIP text encoder and the resulting skill prompt bank remain frozen throughout training, serving as fixed semantic anchors for optimizing the intermediate representations of $\epsilon_\theta$. For the target skill token $S_i$, the skill-aware token sequences are computed according to Eq.~\eqref{encode}, yielding the positive skill-aware sequences $\textbf{v}_{S_i}=[v_{S_i}, v'_{I},v'_{L},v'_{B}]$, which are then fed as the condition for the UNet $\epsilon_\theta$. We then extract the hidden feature representation from the middle layer of the UNet and project it to match the dimensionality of the CLIP features, which produces $h_i$. 

To construct negative pairs, we treat all other skill tokens $S_j \in \mathbb{R}^{D}$ ($j \neq i$) as negative instances which are processed identically to the positive sample to generate mismatched intermediate features $h_j$. The CLIP text embedding of each corresponding incorrect skill category is denoted as $v_{C_j}$. Consequently, the positive pair $(h_i, v_{C_i})$ is contrasted against the negative pairs $(h_j, v_{C_j})$ to enforce semantic alignment in the feature space. We utilize cosine similarity to quantify the alignment
\begin{equation}
    \alpha_{i,j} = \text{cos}(h_i, v_{C_j})
\end{equation}
The skill contrastive loss is formulated as:
\begin{equation}
\mathcal{L}_{\text{Contrastive}} = 
- \log 
\frac{\exp(\alpha_{i,i} / \tau)}
{\exp(\alpha_{i,i} / \tau) + \sum_{j \neq i} \exp(\alpha_{i,j} / \tau)}
\end{equation}
where $\tau$ is a temperature parameter. This objective encourages the model to align the intermediate feature $h_i$ of the correct skill token with its corresponding skill prompt $v_{C_i}$, while suppressing the similarity between the mismatched features $h_j$ induced by incorrect skill tokens and the corresponding incorrect skill prompts $v_{C_j}$. This skill contrastive loss effectively constrains the diffusion model to distinguish between motion patterns of different skills during the denoising process.

\paragraph*{Overall Objective}  
The overall training objective is the weighted combination of the denoising loss and the two skill-constrained objectives:
\begin{equation}
\label{Loss}
    \mathcal{L}_{\text{total}} = \mathcal{L}_{\text{MSE}} + \omega_1 \mathcal{L}_{\text{CE}} + \omega_2 \mathcal{L}_{\text{Contrastive}}
\end{equation}
where $\omega_1$ and $\omega_2$ are hyperparameters that balance the contributions of respective losses. By jointly optimizing these losses, the model learns to predict motion flow that is spatially precise and semantically aligned with the intended skill, thereby achieving robust and skill-aware motion flow generation.

\subsection{Skill-Retrieval Transformation from Flow to Actions}
\label{sec:retrieval}
Mapping the predicted object-centric 2D motion flow $\hat{F} = \{P_1, P_2, \dots, P_{T} \}$ into 3D space and transforming it into executable actions is a critical step. In this section, we first formulate the conventional geometric optimization approach typically used, which serves as the foundation for our method. Subsequently, we introduce our skill-retrieval strategy designed to refine this process.

\subsubsection{Motion Estimation via Geometric Optimization}

Given the initial depth observation captured by the RGB-D camera, the keypoint set $P_1$ in the first frame of $\hat{F}$ is back-projected into 3D space, resulting in $\mathcal{P}_1 \in \mathbb{R}^{N \times 3}$, where $N$ represents the total number of tracked points. To obtain the 3D trajectory of the task-related object, we follow the standard formulation in AVDC~\cite{ko2023actionless}, modeling the rigid motion from the initial frame $t_1$ to any subsequent frame $t$ using a transformation matrix $\boldsymbol{T}_{t_1 \rightarrow t}\in \mathbb{R}^{4\times 4}$. This matrix consists of a rotational component $\boldsymbol{R}_{t_1 \rightarrow t}$ and a translational component $\boldsymbol{t}_{t_1 \rightarrow t}$, describing the object's change in position and orientation within the camera coordinate system. To estimate $\boldsymbol{T}_{t_1 \rightarrow t}$ corresponding to the predicted 2D motion flow $\hat{F}$, we initialize it as an identity matrix and optimize the parameters by minimizing the reprojection error:
\begin{equation}
\label{error}
 E_t = \sum_{n=1}^{N} ||P_{t,n} - K\boldsymbol{T}_{t_1 \rightarrow t}\mathcal{P}_{1,n}||^2
\end{equation}where $E_t$ denotes the optimization objective, $P_{t,n}$ is the $n$-th point in frame $t$ of the predicted flow, $K$ denotes the camera intrinsic matrix, and $\mathcal{P}_{1,n}$ represents the 3D coordinates of the $n$-th point in $\mathcal{P}_1$. By minimizing $E_t$ using a non-linear least squares algorithm, such as Levenberg–Marquardt~\cite{levenberg}, we can obtain the optimized transformation matrix $\boldsymbol{\hat{T}}_{t_1 \rightarrow t}$.

\subsubsection{Skill-Retrieval Transformation Strategy}
While minimizing reprojection error theoretically allows for the mapping of predicted 2D motion flow to 3D trajectories, practical deployments are often hindered by depth ambiguities in subsequent frames and sensor noise, which can lead to trajectory inaccuracies and discontinuities. To address these limitations, we introduce skill-specific trajectory priors into the optimization framework. These priors provide high-level guidance, constraining the optimization process to motion patterns that are consistent with the intended skill, thereby enhancing both precision and physical consistency.

Specifically, we first construct a bank of trajectory templates offline. Utilizing monocular depth estimation (\eg, DepthAnythingV2~\cite{yang2024depth}), we obtain pseudo-depth maps from video demonstrations to lift the extracted 2D motion flow into 3D space. For each skill category, we compute the average 3D trajectory from relevant demonstrations and normalize it spatially. The resulting normalized trajectories form a set of templates $\psi \in \mathbb{R}^{N_s \times T \times 3}$, where $N_s$ denotes the number of skills and $T$ represents the temporal sequence length. During deployment, the skill index predicted by the skill classifier is used to retrieve the corresponding normalized template $\psi_i$. The centers of the task-relevant objects are computed from their 2D bounding boxes and lifted into 3D space using the observed depth. The retrieved template is then translated and scaled according to the initial 3D object positions and their relative distance, yielding a task-specific trajectory prior $\varphi \in \mathbb{R}^{T\times3}$.

We introduce this prior into the optimization process by refining the objective function defined in Eq.~\eqref{error}. The modified objective encourages the estimated 3D trajectory to minimize geometric reprojection error while simultaneously aligning with the retrieved skill prior:
\begin{equation}
\label{error_with_prior}
E_t =
\sum_{n=1}^{N}||P_{t,n}- K\boldsymbol{T}_{t_1 \rightarrow t}\mathcal{P}_{1,n}||^2
+
\lambda
\left|
\bar{\mathcal{P}}_{t} - \varphi_{t}
\right|^2 ,
\tag{16}
\end{equation}
where $\mathcal{P}_{t,n} = \boldsymbol{T}_{t_1 \rightarrow t}\mathcal{P}_{1,n}$ represents the transformed 3D coordinate of the $n$-th keypoint in frame $t$, $\bar{\mathcal{P}}_{t} = \frac{1}{N}\sum_{n=1}^{N}\mathcal{P}_{t,n}$ denotes the centroid of the transformed 3D keypoints, $\varphi_t$ denotes the corresponding object-level waypoint from the skill prior $\varphi$, and $\lambda$ is a hyperparameter balancing the two terms. Thus, the prior term constrains the object-level translational trajectory while preserving the rigid structure among all keypoints. By optimizing this joint objective, the system produces a transformation matrix $\boldsymbol{\hat{T}}_{t_1 \rightarrow t}$ that is not only geometrically plausible but also robust to noise and faithful to the characteristic motion patterns of the skill. The complete pipeline of the skill-retrieval transformation strategy is summarized in Algorithm~\ref{alg:skill_retrieval}.

\subsubsection{Mapping to Executable Actions}
The optimized transformation matrix is further transformed from the camera coordinate system to the robot base coordinate system using a camera-to-base transformation obtained through standard eye-to-hand calibration. By applying the relative offset between the end-effector and the object, we derive the target 3D end-effector path. Inverse kinematics (IK) is then applied to generate smooth joint position and velocity commands, completing the mapping from 2D flow to executable actions.

\begin{algorithm}[t]
\caption{Skill-Retrieval Transformation Strategy}
\label{alg:skill_retrieval}
\SetKwInOut{Input}{Input}
\SetKwInOut{Output}{Output}
\Input{Predicted 2D motion flow $\hat{F} = \{P_1, \dots, P_T\}$, initial depth frame, camera intrinsics $K$, skill templates $\psi$, skill category $s$}
\Output{Optimized 3D transformation matrix $\{\boldsymbol{\hat{T}}_{t_1 \rightarrow t}\}_{t=1}^{T}$}
Back-project keypoints $P_1$ using initial depth to obtain initial 3D points $\mathcal{P}_1 \in \mathbb{R}^{N \times 3}$\;
Retrieve skill template $\psi_s \in \mathbb{R}^{T \times 3}$ corresponding to skill category $s$\;
Align $\psi_s \in \mathbb{R}^{T \times 3}$ with initial object configuration to obtain trajectory prior $\varphi \in \mathbb{R}^{T \times 3}$\;
\For{$t = 1$ \KwTo $T$}{
    Initialize transformation $\boldsymbol{T}_{t_1 \rightarrow t} \gets \mathbf{I}_{4 \times 4}$\;
    Define joint optimization objective $E_t$ by Eq.~\eqref{error_with_prior}\;
    Optimize $E_t$ using a non-linear least
    squares algorithm (\eg, Levenberg–Marquardt) to obtain $\boldsymbol{\hat{T}}_{t_1 \rightarrow t}$\;
}
\Return{$\{\boldsymbol{\hat{T}}_{t_1 \rightarrow t}\}_{t=1}^{T}$}
\end{algorithm}

\section{Experiments}
\label{sec:experiments}

In this section, we first introduce the dataset construction in Section~\ref{sec:dataset}, followed by the training details in Section~\ref{sec:details}. Section~\ref{sec:sim_results} presents comparative evaluations across within-distribution, challenging generalization, and skill adaptation settings, alongside component-wise ablation studies in simulation. Section~\ref{sec:real_results} verifies the method's generalization and zero-shot sim-to-real transfer capabilities in real-world scenarios. Finally, Section~\ref{sec:scale_comp} evaluates the scalability and composability of the proposed approach.

\begin{figure*}[t]
    \centering
    \includegraphics[width=1.0\textwidth]{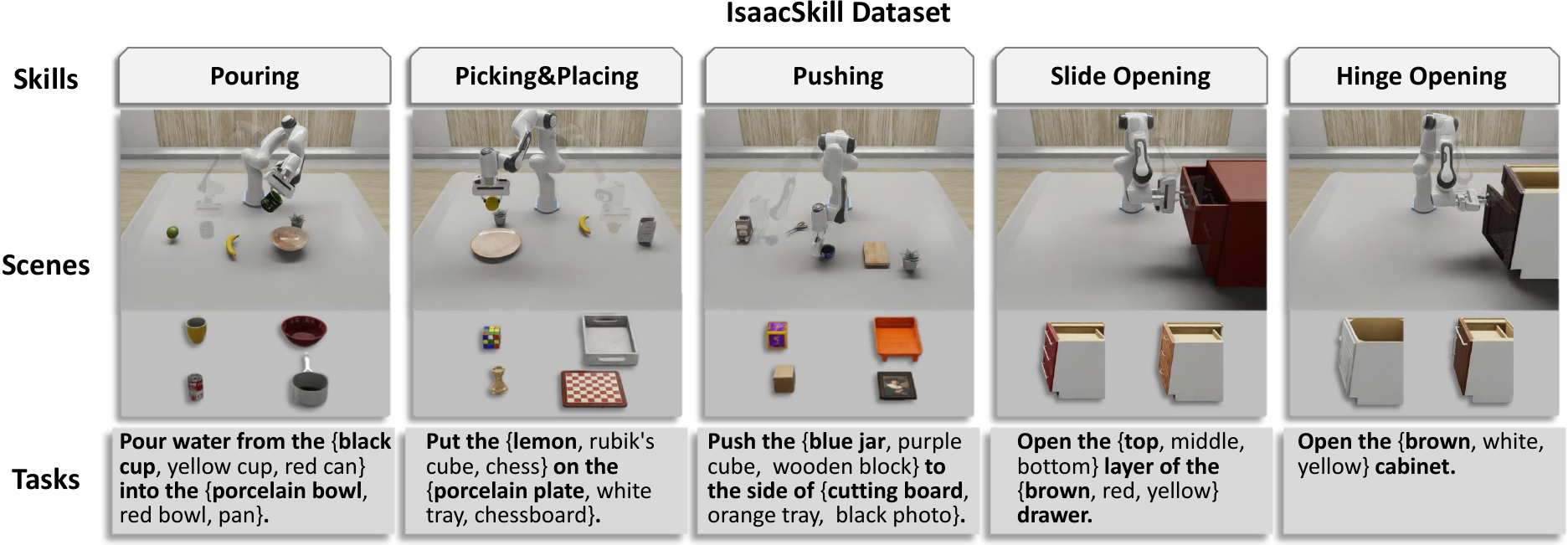}
    \caption{Overview of the constructed IsaacSkill dataset. The dataset comprises five fundamental manipulation skills: ``\textit{Pouring}'', ``\textit{Picking \& Placing}'',  ``\textit{Pushing}'', ``\textit{Slide Opening}'', and ``\textit{Hinge Opening}''. Each skill includes three different tasks involving different objects.}
    \label{fig_sim_all_tasks}
\end{figure*}

\subsection{Dataset Construction}
\label{sec:dataset}

While datasets such as Meta-World~\cite{yu2020meta}, RLBench~\cite{james2020rlbench}, and LIBERO~\cite{liu2023libero} have advanced the field of robotic manipulation, their simulation environments have limitations in physical and visual fidelity. These limitations can make it difficult to fully capture the dynamics involved in real-world interactions, posing challenges for evaluating generalization under realistic conditions and achieving zero-shot sim-to-real transfer. To bridge this fidelity gap, we constructed IsaacSkill, a novel dataset built on the high-fidelity NVIDIA Isaac Lab~\cite{isaaclab} platform, ensuring physically accurate dynamics and photorealistic scenarios. Furthermore, unlike prior works~\cite{im2flow2act,track2act,ko2023actionless} that focus on a narrow set of isolated tasks, IsaacSkill is structured around 5 fundamental robotic skills: ``\textit{Pouring}'', ``\textit{Picking \& Placing}'', ``\textit{Pushing}'', ``\textit{Slide Opening}'', and ``\textit{Hinge Opening}''. This skill-centric approach allows us to encompass a wide variety of tasks involving diverse objects and scenarios, as shown in  Fig.~\ref{fig_sim_all_tasks}. To ensure robustness, we design 3 different tasks per skill, introducing high variability in object geometry, texture, and placement, alongside randomized distractor objects. In total, the dataset comprises 2,400 trajectory demonstrations (160 per task), providing a comprehensive benchmark for evaluating generalization across diverse tasks within each skill category.

In our dataset, the robotic platform consists of a Franka Emika Panda manipulator equipped with a parallel-jaw gripper. A single RGB camera is positioned at an elevated frontal angle to provide a third-person view of the entire workspace. During data collection, we employ an oracle policy to ensure successful execution for each task. The resulting demonstrations are recorded as single-view RGB videos and temporally downsampled to 32 frames to maintain consistency. To extract the motion flow of task-relevant objects, we leverage Qwen-VL~\cite{Qwen-VL} to first localize target objects based on language instructions. Subsequently, keypoints are uniformly sampled within the detected bounding boxes and tracked throughout the video sequence using the TAPIR tracker~\cite{doersch2023tapir}.

\subsection{Training Details}
\label{sec:details}

The training process is decoupled into two stages. In the first stage, the encoder $E_\phi(\cdot)$ and the decoder $D_\phi(\cdot)$ are trained within an autoencoder framework. We initialize both components using pretrained Stable Diffusion~\cite{stablediffusion} weights, but keep the encoder frozen while fine-tuning the decoder on our dataset. In the second stage, the learnable skill tokens, the skill classifier and the noise prediction network $\epsilon_\theta$ are jointly trained. The parameters of $\epsilon_{\theta}$ are initialized from Stable Diffusion~\cite{stablediffusion}, except for the motion module, which is trained from scratch. The remaining layers are fine-tuned using Low-Rank Adaptation~\cite{lora} (LoRA) to ensure parameter efficiency. We employ the AdamW optimizer with an initial learning rate of $1 \times 10^{-5}$, utilizing a cosine decay schedule. The hyperparameters in Eq.~\eqref{Loss} are set to $\omega_1=0.01$ and $\omega_2=0.02$, based on a sequential hyperparameter search conducted on the validation set using average displacement error (ADE) and final displacement error (FDE) as the evaluation metrics. The number of denoising steps $m$ in the diffusion process is set to 40. All training is conducted on four NVIDIA A100 GPUs, where the first and second stages require approximately 7.2 hours and 44.1 hours, respectively.

\subsection{Simulation Experiments}
\label{sec:sim_results}
\subsubsection{Experimental Setup}
We conduct simulation experiments on the proposed IsaacSkill dataset. Four baseline methods representing diverse paradigms are selected for comparison:
\begin{itemize}
\item \emph{R3M}~\cite{nair2023r3m}: A behavior cloning approach utilizing pretrained visual representations. It leverages temporal contrastive learning and video-language alignment to extract semantically rich features for robust policy learning.
\item \emph{AVDC}~\cite{ko2023actionless}: A video prediction framework that synthesizes task-specific future frames and derives executable actions by analyzing dense pixel correspondences between adjacent predicted frames.
\item \emph{Track2Act}~\cite{track2act}: A trajectory-centric method that employs a goal-conditioned grid flow prediction model to generate future trajectories for query points. These trajectories are subsequently mapped to residual actions via a specific policy network.
\item \emph{Im2Flow2Act}~\cite{im2flow2act}: A flow-based approach that generates object-centric motion flow that serves as a condition for a policy network to output executable actions. For a fair comparison, we replace Grounding DINO~\cite{liu2024grounding} with Qwen-VL~\cite{Qwen-VL} to locate task-relevant objects.
\end{itemize}

To comprehensively evaluate the effectiveness and robustness of the proposed SADiff, we structure our simulation experiments into four main parts:
\begin{itemize}
    \item \emph{Within-Distribution Comparative Experiment}: We first assess the fundamental imitation capabilities of SADiff and baselines on tasks consistent with the training distribution.
    \item \emph{Robustness and Generalization Analysis}: We rigorously test the model's robustness against variations in visual context, object instances, categories, and robotic embodiments to evaluate out-of-distribution performance.
    \item \emph{Instruction-Guided Skill Adaptation Experiment}: We examine the model's flexibility in adapting to changing language instructions while the scene configuration remains constant.
    \item \emph{Ablation Studies}: Finally, we isolate and quantify the contributions of key components, including learnable skill tokens, skill contrastive loss, and skill-specific trajectory priors.
\end{itemize}

In all the aforementioned experiments, we compare our method against baselines over 25 independent rollouts, reporting the average success rate (SR) as the primary metric. To ensure robust evaluation, each rollout uses randomly initialized positions for task-related objects, as well as randomized types and placements for interfering objects. In the following sections, we present detailed results and analysis for each experimental setting.

\begin{figure*}[t]
    \centering
    \includegraphics[width=1.0\linewidth]{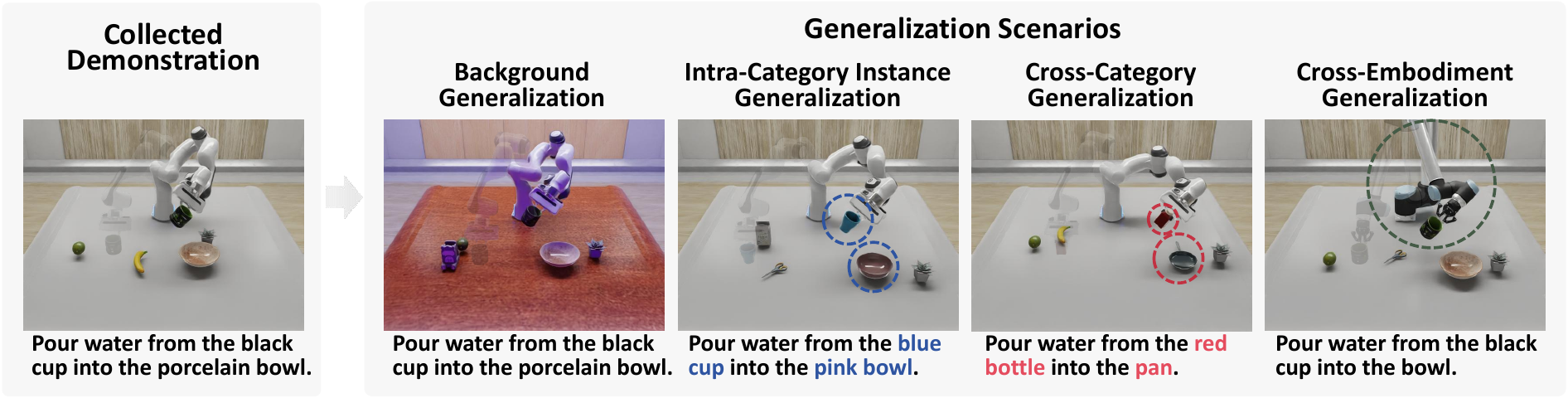}
    \caption{Overview of the generalization evaluation settings. Left: The original collected demonstration. Right: The model is evaluated under four different types of variations involving distribution shifts in backgrounds, intra-category instances, cross-category objects, and embodiment.}
    \label{fig_ger}
\end{figure*}

\subsubsection{Within-Distribution Comparative Experiment}
To evaluate the fundamental imitation capability of the proposed method, we conduct a standard within-distribution comparative experiment, serving as a primary baseline for imitation performance. The scenes and task-related objects are sampled from exactly the same distributions as the collected dataset, ensuring a rigorous assessment of the system's ability to recall and execute demonstrated skills.

\begin{table}[t]
\caption{Comparison of Success Rates (\%) Across Different Skills in Within-Distribution Comparative Experiment}
\label{tab:performance_contra_sim}
\renewcommand{\arraystretch}{1.2}
\setlength{\tabcolsep}{1.15pt}
\begin{tabular}{l|cccccc}
\toprule
\textbf{Method} & \textbf{Pouring} & \textbf{\makecell{Picking\&\\Placing}} & \textbf{Pushing} & \textbf{\makecell{Slide\\Opening}} & \textbf{\makecell{Hinge\\Opening}} & \textbf{\makecell{Average\\SR}} \\
\midrule
R3M~\cite{nair2023r3m} & 64.0 & 64.0 & 52.0 & 60.0 & 60.0 & 60.0 \\
AVDC~\cite{ko2023actionless} & 64.0 & 68.0 & 84.0 & 92.0 & 72.0 & 76.0 \\
Track2Act~\cite{track2act} & 88.0 & 80.0 & 76.0 & 80.0 & 88.0 & 82.4 \\
Im2Flow2Act~\cite{im2flow2act} & 84.0 & 92.0 & 88.0 & 92.0 & 84.0 & 88.0 \\
\textbf{SADiff (Ours)} & \textbf{92.0} & \textbf{96.0} & \textbf{92.0} & \textbf{96.0} & \textbf{88.0} & \textbf{92.8} \\
\bottomrule
\end{tabular}
\end{table}

The experimental results are presented in Table~\ref{tab:performance_contra_sim}. SADiff consistently outperforms all baselines across the five skills, achieving a state-of-the-art average success rate of 92.8\%. Notably, SADiff surpasses the closest competitor, Im2Flow2Act (88.0\%), by 4.8 percentage points. Since Im2Flow2Act also leverages object-centric motion flow, this performance gain suggests that explicitly modeling skill-level information helps improve the quality of generated motion flow and the accuracy of action transformation. Conversely, methods modeling global scene dynamics, including Track2Act (82.4\%) and AVDC (76.0\%), achieve lower success rates because their full-frame synthesis or dense grid tracking increases susceptibility to background perturbations. SADiff addresses this limitation by focusing exclusively on task-relevant objects, effectively filtering out visual noise. Moreover, R3M (60.0\%) yields the lowest success rate, indicating that pretrained general-purpose visual representations struggle to disentangle fine-grained, task-specific dynamics from environmental distractors. Collectively, these results confirm that integrating skill-aware priors with focused object dynamics is essential for robust manipulation in within-distribution settings.

\subsubsection{Robustness and Generalization Analysis}
\label{sec:robustness}

To rigorously evaluate the robustness and generalization of SADiff, we design experiments across diverse scenarios that deviate significantly from the training distribution. Specifically, we investigated the performance under four different types of variations involving backgrounds, object instances, object categories, and robotic embodiments. As illustrated in Fig.~\ref{fig_ger}, the four generalization settings are designed as follows:
\begin{itemize}
  \item \emph{Background Generalization}: This setting introduces variations in lighting conditions and background while keeping the task-related objects consistent with the training set.
  \item \emph{Intra-Category Instance Generalization}: This setting evaluates the model on unseen instances of task-related objects that differ in color, shape, and physical appearance within the same category.
  \item \emph{Cross-Category Generalization}: This setting challenges the model by replacing task-related objects with functionally similar objects from entirely different categories (\eg, replacing a cup with a bottle).
  \item \emph{Cross-Embodiment Generalization}: This setting assesses cross-embodiment generalization by replacing the Franka Panda manipulator with a UR10 robotic arm equipped with a Robotiq two-finger gripper.
\end{itemize}

\begin{figure}[t]
    \centering
    \includegraphics[width=1\linewidth]{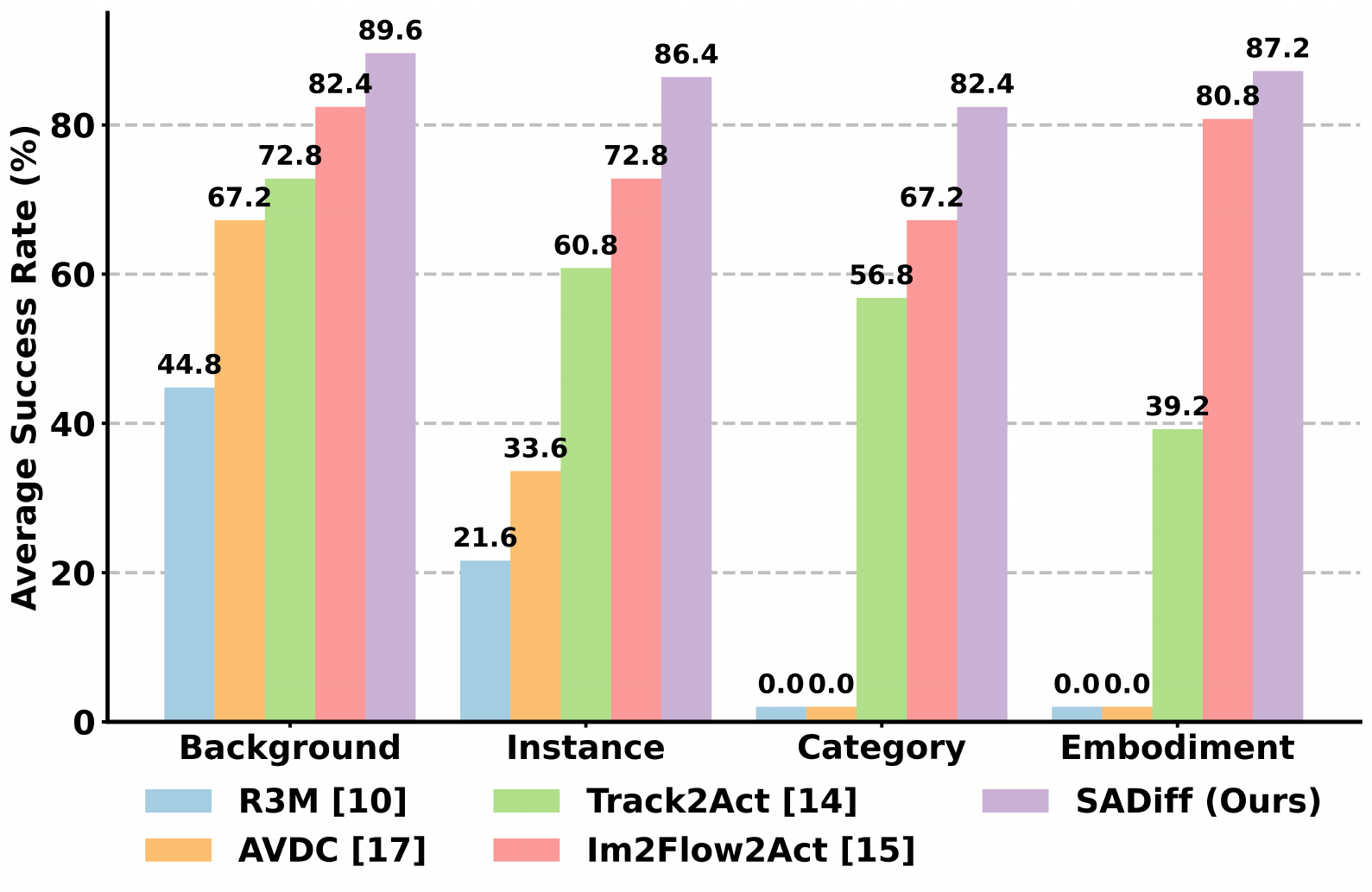}
    \caption{Average success rate (\%) of baselines and SADiff under different generalization scenarios. SADiff consistently achieves the highest average success rate across all four generalization settings.}
    \label{fig_result}
\end{figure}

The quantitative results averaged across five skills in Fig.~\ref{fig_result} demonstrate that SADiff consistently surpasses all baselines across every generalization setting. In the background and intra-category instance generalization scenarios, SADiff achieves robust average success rates of 89.6\% and 86.4\% by leveraging skill-level guidance to overcome visual variations. This mechanism prevents the incoherent motion flow observed in Im2Flow2Act and mitigates the visual distractions that degrade the performance of R3M, AVDC, and Track2Act. In the cross-category generalization setting, SADiff maintains an 82.4\% success rate while AVDC and R3M fail completely and other baselines suffer performance drops exceeding 20\%. These results indicate that skill priors effectively bridge the gap between different object categories. Furthermore, the cross-embodiment setting reveals that R3M fails due to proprioception mismatches and AVDC cannot synthesize plausible future frames for the unseen robot. While Track2Act and Im2Flow2Act also suffer significant degradation due to limited adaptability, SADiff exhibits a decrease of only 5.6 percentage points in success rate. This stability arises from the skill-specific trajectory priors that enable effective adaptation to novel robotic platforms. Overall, these experiments validate that the skill-aware design of SADiff is essential for robust generalization across various visual, object, and embodiment variations.

\subsubsection{Instruction-Guided Skill Adaptation Experiment}
\begin{figure}[t]
    \centering
    \includegraphics[width=1.0\linewidth]{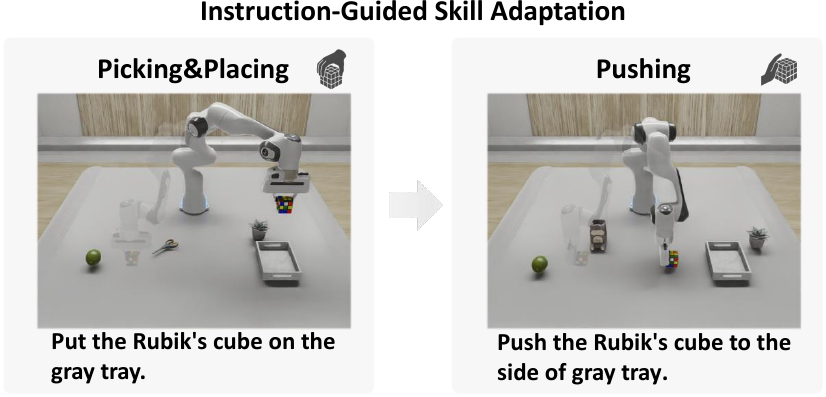}
    \caption{Examples of the instruction-guided adaptation scenario. The model is evaluated on its ability to switch manipulation skills within an unchanged visual environment solely by following modified language instructions.}
    \label{fig_cross_skill}
\end{figure}

To further investigate whether the proposed SADiff can overcome spurious scene-task correlations and prioritize semantic instructions over visual priors, we design an instruction-guided adaptation experiment. In this setup, the physical scene configuration and object placements remain consistent with the training distribution, but the language instruction is altered to command a different manipulation skill. As illustrated in Fig.~\ref{fig_cross_skill}, a scene originally associated with a ``\textit{Picking \& Placing}'' task is redefined as a ``\textit{Pushing}'' task during testing. This protocol rigorously evaluates whether the model can flexibly adapt its behavior to new semantic instructions rather than simply memorizing scene-task correlations. We define four specific cross-skill adaptation tasks:

\begin{itemize}
    \item \emph{Task 1}: ``\textit{Picking \& Placing}'' $\rightarrow$ ``\textit{Pushing}''
    \item \emph{Task 2}: ``\textit{Pushing}'' $\rightarrow$ ``\textit{Picking \& Placing}''  
    \item \emph{Task 3}: ``\textit{Pouring}'' $\rightarrow$ ``\textit{Picking \& Placing}''  
    \item \emph{Task 4}: ``\textit{Pouring}'' $\rightarrow$ ``\textit{Pushing}''  
\end{itemize}

The experimental results in Table~\ref{tab:performance_taskchange_sim} highlight the superior flexibility of SADiff, which achieves an average success rate of 85.0\% and significantly outperforms all baselines. R3M and AVDC exhibit a complete inability to adapt with 0\% success rates, because they rigidly rely on the specific scene-task correlations learned during training. Similarly, Im2Flow2Act performs poorly with only 16.0\% success as its flow generation is overly constrained by the visual scene rather than the language instruction. Although Track2Act fares slightly better with 41.0\% due to goal-image conditioning, it still lacks the semantic understanding necessary to consistently execute the new skill. SADiff successfully overcomes these limitations by explicitly classifying the modified instruction into a known skill category during inference. This mechanism allows the model to leverage learned skill representations and generate appropriate actions that align with the new command even when the visual environment suggests a different task.

\begin{table}[t]
\centering\small
\caption{Comparison of Success Rates (\%) of Instruction-Guided Skill Adaptation Experiment}
\label{tab:performance_taskchange_sim}
\renewcommand{\arraystretch}{1.2}
\setlength{\tabcolsep}{2.5pt}
\begin{tabular}{l|ccccc}
\toprule
\textbf{Method} & \textbf{Task 1} & \textbf{Task 2} & \textbf{Task 3} & \textbf{Task 4} & \textbf{Average SR} \\
\midrule
R3M~\cite{nair2023r3m} & 0.0 & 0.0 & 0.0 & 0.0 & 0.0 \\
AVDC~\cite{ko2023actionless} & 0.0 & 0.0 & 0.0 & 0.0 & 0.0 \\
Track2Act~\cite{track2act} & 44.0 & 52.0 & 32.0 & 36.0 & 41.0 \\
Im2Flow2Act~\cite{im2flow2act} & 20.0 & 16.0 & 12.0 & 16.0 & 16.0 \\
\textbf{SADiff (Ours)} & \textbf{88.0} & \textbf{84.0} & \textbf{88.0} & \textbf{80.0} & \textbf{85.0} \\
\bottomrule
\end{tabular}
\end{table}

\subsubsection{Ablation Studies}
To systematically assess the individual contributions of each core component, we perform ablation experiments following the same evaluation protocol utilized for the Robustness and Generalization Analysis in Section~\ref{sec:robustness}. Specifically, we evaluate performance across background, intra-category instance, and cross-category generalization settings. The three ablated variants are defined as follows:
\begin{itemize}
\item \emph{w/o Learnable Skill Tokens (LST)}: This variant removes the learnable skill tokens, along with the related skill classification loss and skill contrastive loss.
\item \emph{w/o Skill Contrastive Loss (SCL)}: In this variant, the skill contrastive loss in Eq.~\eqref{Loss} is removed.
\item \emph{w/o Skill Trajectory Priors (STP)}: In this variant, the skill-specific trajectory prior in Eq.~\eqref{error_with_prior} is disabled.
\end{itemize}

The results are shown in Table~\ref{tab:performance_ablation}. Removing the learnable skill tokens (LST)  causes the average success rate to decline from 86.1\% to 75.7\%. This drop indicates that without explicit skill representations, the model loses the ability to dynamically modulate multimodal inputs with skill-specific context, severely hampering adaptability in unseen scenarios. Removing the skill contrastive loss (SCL) reduces the average success rate to 81.1\%, suggesting that without skill-level constraints during the denoising process, the diffusion model generates motion flow that drift from the intended manipulation semantics. Most critically, removing the skill trajectory priors (STP) causes the largest performance drop, reducing the average success rate to 66.9\%. This highlights that trajectory priors are essential for stabilizing the transformation from 2D motion flow to 3D executable actions. Collectively, these findings demonstrate that SADiff relies on the tight coupling of skill-aware encoding, constrained diffusion, and prior-guided transformation to achieve robust generalization.

\begin{figure*}[t]
    \centering
    \includegraphics[width=1.0\textwidth]{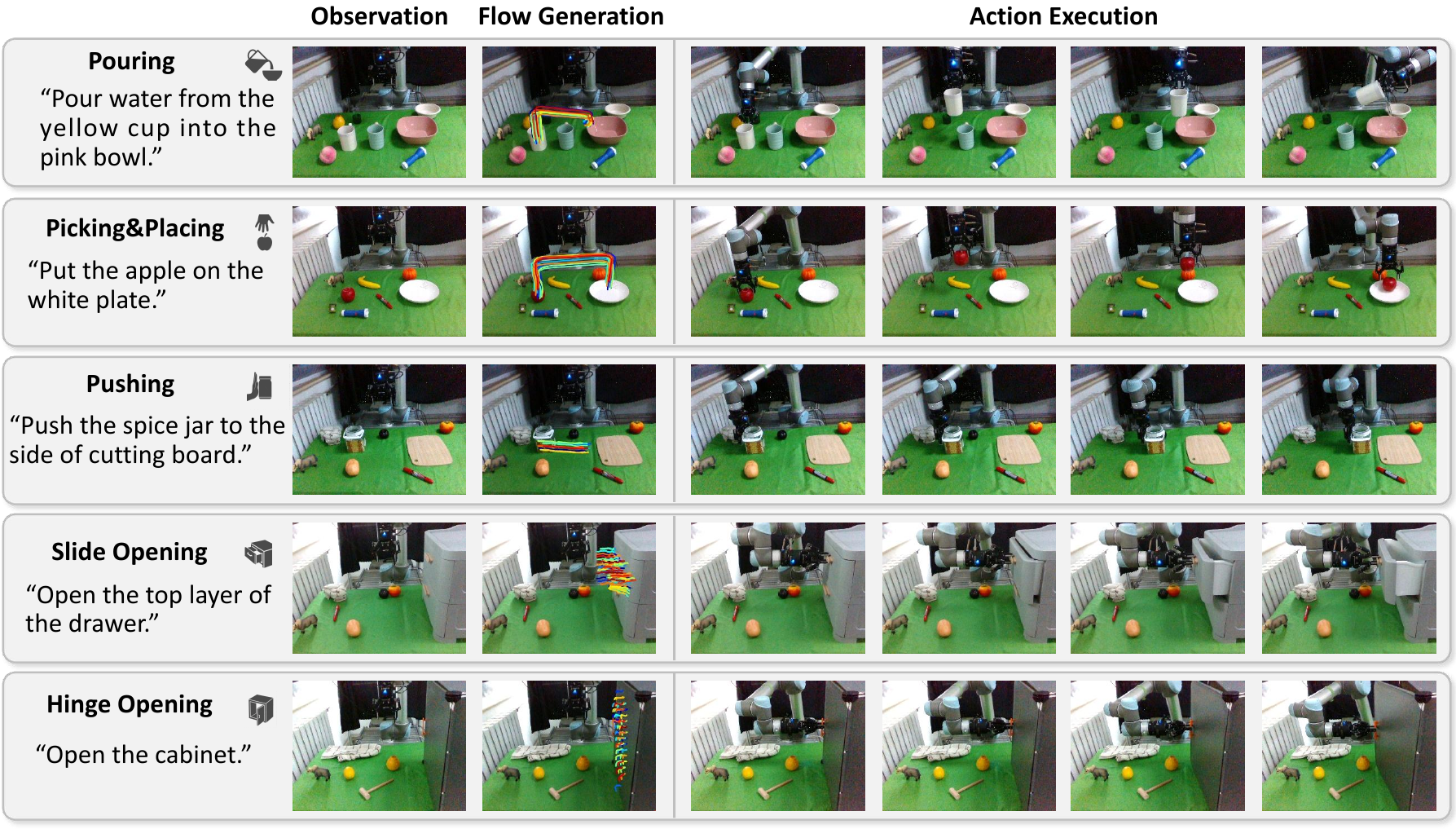}
    \caption{Visualization of real-world experiments. This figure demonstrates the predicted object-centric motion flow and the corresponding execution trajectories for five manipulation skills in the real world.}
    \label{fig_real}
\end{figure*}

\begin{table}[t]
    \centering
    \caption{Comparison of Success Rates (\%) of Ablation Studies}
    \label{tab:performance_ablation}
    \renewcommand{\arraystretch}{1.2}
    \setlength{\tabcolsep}{3.7pt}
    \begin{tabular}{l|cccc}
        \toprule
        \textbf{Method} & \textbf{\makecell{Background\\Generalization}} & \textbf{\makecell{Instance\\Generalization}} & \textbf{\makecell{Cross-Category\\Generalization}} &  \textbf{\makecell{Average\\SR}} \\
        \midrule
        \textbf{SADiff} & \textbf{89.6} & \textbf{86.4} & \textbf{82.4} & \textbf{86.1} \\
        w/o LST & 84.8 & 74.4 & 68.0 & 75.7 \\
        w/o SCL & 86.4 & 81.6 & 75.2 & 81.1 \\
        w/o STP & 73.6 & 62.4 & 64.8 & 66.9 \\
        \bottomrule
    \end{tabular}
\end{table}

\subsection{Real-World Experiments}
\label{sec:real_results}

\subsubsection{Experimental Setup}
In real-world experiments, we employ a UR5 robotic manipulator equipped with a Robotiq gripper and an Intel RealSense D435i camera, which is mounted above the front side of the workspace to maintain consistency with the simulation setup. The model is deployed for inference on a workstation equipped with a single NVIDIA RTX 4090 GPU and an Intel Core i9-13900KF CPU. To evaluate the generalization ability and zero-shot sim-to-real transfer capability of SADiff, we directly deploy the model trained purely in simulation without using real-world demonstrations, task-specific fine-tuning, or task-dependent parameter adjustment. The only additional setup is an eye-to-hand calibration to estimate the transformation between the camera and robot base coordinate frames. We compare our approach against Track2Act~\cite{track2act} and Im2Flow2Act~\cite{im2flow2act}, as other baselines such as R3M~\cite{nair2023r3m} and AVDC~\cite{ko2023actionless} lack the capability for direct transfer. For each skill, we conduct 25 independent rollouts. To rigorously test robustness, we follow the simulation setting by randomly initializing the positions of task-related objects and introducing randomized types and placements for visual distractors in each trial.

\begin{figure*}[t]
    \centering
    \includegraphics[width=1.0\textwidth]{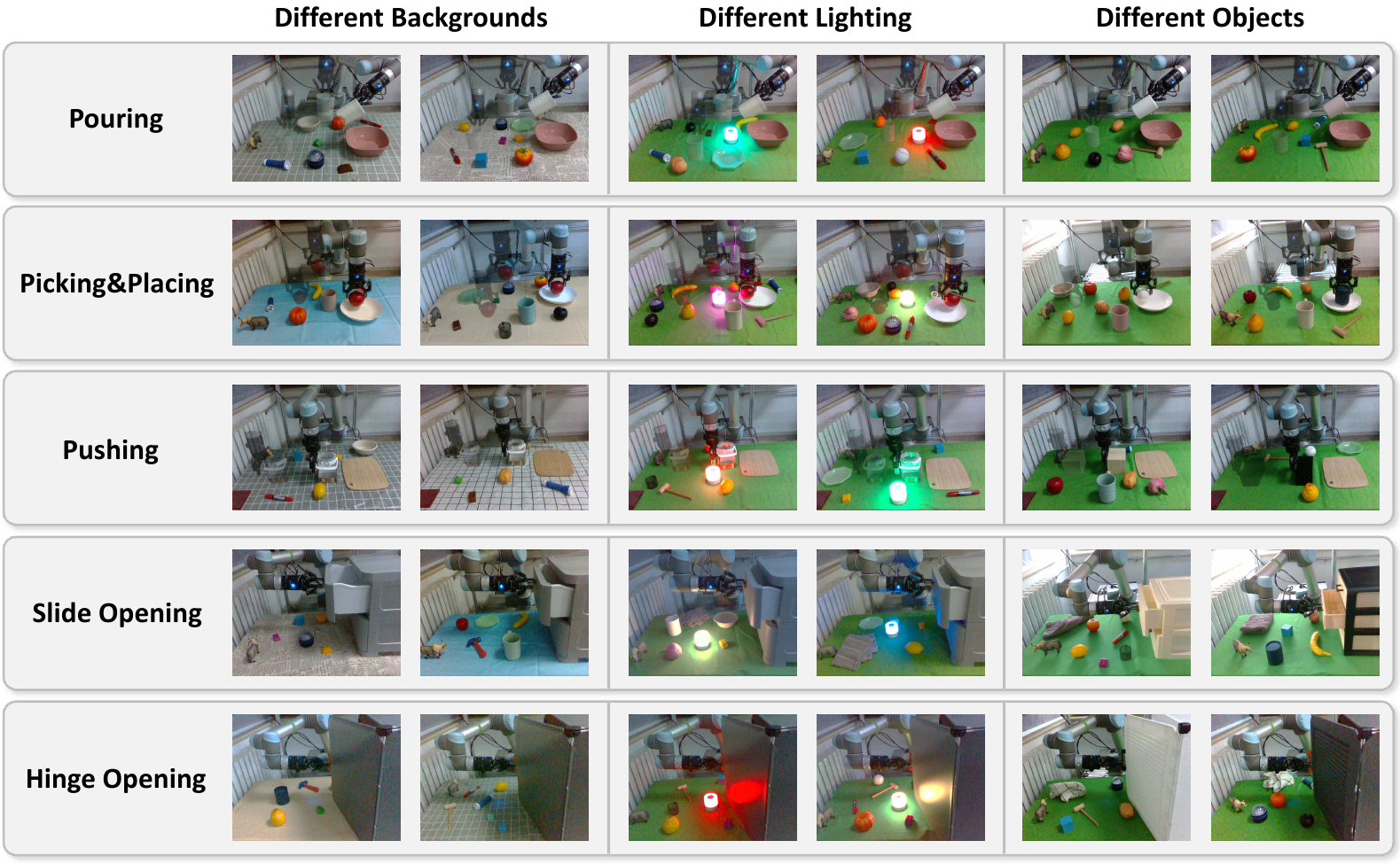}
    \caption{Qualitative evaluation under environmental variations. We subject the model to a variety of test scenarios featuring diverse backgrounds, complex lighting conditions, and unseen object instances to assess its visual robustness.}
    \label{fig_var}
\end{figure*}

\subsubsection{Experiment Results}
The results are summarized in Table~\ref{tab:performance_real}. SADiff achieves an average success rate of 76.0\% across all five skills, exceeding those of Im2Flow2Act and Track2Act by 21.6 and 25.6 percentage points, respectively. The primary challenge to achieving robust zero-shot sim-to-real transfer lies in pervasive domain discrepancies between simulation and physical environments. These discrepancies include substantial changes in visual scenes and illumination conditions, sensor noise, variations in object types, and embodiment differences. These factors typically cause severe performance degradation for policies trained exclusively in simulation. Baseline methods struggle considerably under these conditions. Although Track2Act can generate query-point trajectories conditioned on goal images, it is highly sensitive to visual domain shifts, which degrades trajectory quality. Furthermore, its residual policy, which was trained to map 2D trajectories to actions for a specific simulated agent, generalizes poorly to the physical robot. Similarly, Im2Flow2Act leverages a VLM to identify task-relevant objects, yet it falters when generating object-centric motion flow for novel scenes or unseen instructions. Crucially, its learned policy for converting 2D flow into executable actions relies heavily on fixed camera parameters, limiting its adaptability to real-world setups.

\begin{table}[t]
\caption{Comparison of Success Rates (\%) Across Different Skills in Real-World Experiment}
\label{tab:performance_real}
\renewcommand{\arraystretch}{1.2}
\setlength{\tabcolsep}{1.15pt}
\begin{tabular}{l|cccccc}
\toprule
\textbf{Method} & \textbf{Pouring} & \textbf{\makecell{Picking\&\\Placing}} & \textbf{Pushing} & \textbf{\makecell{Slide\\Opening}} & \textbf{\makecell{Hinge\\Opening}} & \textbf{\makecell{Average\\SR}} \\
\midrule
Track2Act~\cite{track2act} & 44.0 & 52.0 & 56.0 & 48.0 & 52.0 & 50.4 \\
Im2Flow2Act~\cite{im2flow2act} & 56.0 & 60.0 & 56.0 & 60.0 & 40.0 & 54.4 \\
\textbf{SADiff (Ours)} & \textbf{72.0} & \textbf{80.0} & \textbf{80.0} & \textbf{76.0} & \textbf{72.0} & \textbf{76.0} \\
\bottomrule
\end{tabular}
\end{table}

In contrast, SADiff effectively bridges the sim-to-real gap by injecting skill-level knowledge into both the flow generation and action transformation stages. As demonstrated by the real-world results, SADiff retains strong performance without requiring domain randomization or fine-tuning. This consistent superiority validates two key advantages: (1) explicit skill modeling stabilizes diffusion-based motion generation against visual noise, and (2) the skill-retrieval transformation module reliably converts predicted motion flow into executable actions, even across different robotic embodiments. These findings confirm that SADiff not only excels in simulation but also robustly handles embodiment changes and physical variations in real-world environments. Fig.~\ref{fig_real} visualizes the predicted object-centric motion flow and corresponding execution trajectories for five skills, providing further qualitative evidence of the model's robustness.

In addition, we investigated the time efficiency of the proposed framework in real-world deployment. We measured the average planning times for flow generation and 2D-to-3D mapping, as well as the average execution time, as shown in Table~\ref{tab:timecost}. The overall planning latency is primarily dominated by VLM inference and the iterative sampling process of the diffusion model. In comparison, skill-aware encoding introduces relatively little computational overhead because its attention operations are performed on compact token sequences only once per task, while skill-retrieval transformation mainly involves depth back-projection, prior alignment, and a small-scale geometric optimization. The reported planning times, ranging from 16.9 to 23.1 seconds, correspond to task-level planning performed once before trajectory execution, rather than high-frequency closed-loop replanning. Therefore, the current implementation is suitable for manipulation tasks that permit task-level planning latency, but it does not yet support real-time replanning at the control frequency.

\begin{table}[t]
\caption{Planning and Execution Time for Five Skills in Real-World Experiments}
\label{tab:timecost}
\renewcommand{\arraystretch}{1.2}
\setlength{\tabcolsep}{3.15pt}
\begin{tabular}{l|c c c c c}
\toprule
\textbf{Phases} & \textbf{Pouring} & \textbf{\makecell{Picking\&\\Placing}} & \textbf{Pushing} & \textbf{\makecell{Slide\\Opening}} & \textbf{\makecell{Hinge\\Opening}} \\
\midrule 
\textbf{Planning Time (s)} & 18.4 & 23.1 & 17.2 & 16.9 & 17.8 \\
\textbf{Execution Time (s)} & 28.2 & 21.9 & 10.6 & 14.5 & 15.2 \\
\bottomrule
\end{tabular}
\end{table}

\begin{figure*}[t]
    \centering
    \includegraphics[width=1.0\textwidth]{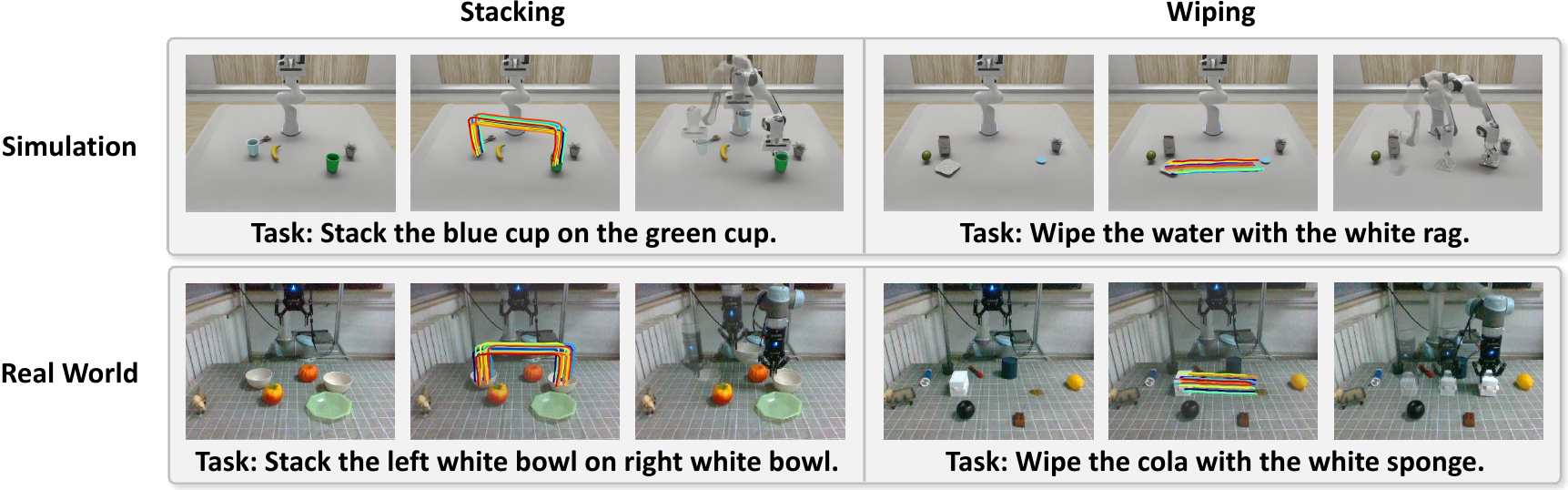}
    \caption{Visualization of the scalability of SADiff to unseen skills. We show that SADiff can successfully perform new skills (e.g., ``\textit{Stacking}'' and ``\textit{Wiping}'') unseen during training in both simulated and real-world environments.}
    \label{scalability_fig}
\end{figure*}

\subsubsection{Qualitative Analysis under Environmental Variations}
Beyond the quantitative metrics, we qualitatively evaluated how the policy behaves under extreme visual distractions and domain discrepancies. We extended the experimental evaluation to include real-world scenarios characterized by significant environmental variations. As illustrated in Fig.~\ref{fig_var}, we introduced three different types of domain shifts to evaluate the model's visual robustness:
\begin{itemize}
    \item \emph{Different Backgrounds}: We altered the workspace environment using tablecloths with different textures and colors to test robustness against background clutter.
    \item \emph{Different Lighting}: We introduced complex lighting conditions, including drastic changes in color and intensity, to evaluate the model's robustness to illumination shifts.
    \item \emph{Different Objects}: We replaced the manipulation targets with unseen object instances that differ in shape, size, and appearance.
\end{itemize}

We constructed a diverse evaluation set comprising 30 different tasks spanning all five manipulation skills. Despite domain shifts and visual distractions, SADiff demonstrates remarkable adaptability. The qualitative results indicate that our method effectively disentangles target object dynamics from environmental noise. This resilience is attributed to the combination of object-centric motion flow and the structural constraints imposed by skill-specific priors, which enables SADiff to perform reliably even in highly unstructured and visually novel environments. These findings further validate the potential of SADiff for robust open-world deployment.

\subsection{Scalability and Composability Experiments}
\label{sec:scale_comp}
To bridge the gap between specialized manipulation policies and general-purpose robotic agents, it is crucial to verify capabilities beyond standard task repetition. Specifically, real-world deployment requires an agent not only to generalize to behaviors not seen during training but also to seamlessly sequence mastered skills to solve complex, multi-stage problems. To further validate the scalability and robustness of the proposed SADiff in these advanced settings, we consider two essential properties: (1) scalability to unseen skills absent from the training dataset, demonstrating the transferability of our skill representations; and (2) composability for completing long-horizon tasks by chaining multiple skill executions. All experiments use the SADiff model pretrained on IsaacSkill without additional fine-tuning and evaluate it in both simulation and real-world environments.

\subsubsection{Scalability to Unseen Skills}
\label{sec:scalability}
We validate scalability by introducing two unseen skills, ``\textit{Stacking}'' and ``\textit{Wiping}'', which were entirely absent from the training phase. As visualized in Fig.~\ref{scalability_fig}, SADiff successfully generates coherent and temporally consistent motion flow for both tasks, enabling reliable translation into stable action trajectories. This zero-shot generalization is achievable because the fundamental skills in our training set (\eg, \textit{``Picking \& Placing'' and ``Pushing''}) share underlying motion patterns with these novel skills. Consequently, SADiff can effectively approximate the required behaviors for unseen skills like ``\textit{Stacking}'' and ``\textit{Wiping}'' by leveraging the learned primitives from these foundational skills. These results demonstrate that the proposed SADiff framework can seamlessly extend to semantically and kinematically related behaviors without additional training. Furthermore, this experiment suggests that as the number and diversity of fundamental skills in IsaacSkill increase, the model's capacity to master complex and diverse tasks will scale correspondingly.

\subsubsection{Composability for Long-Horizon Tasks}
\label{sec:composability}
\begin{figure*}[t]
    \centering
    \includegraphics[width=1.0\textwidth]{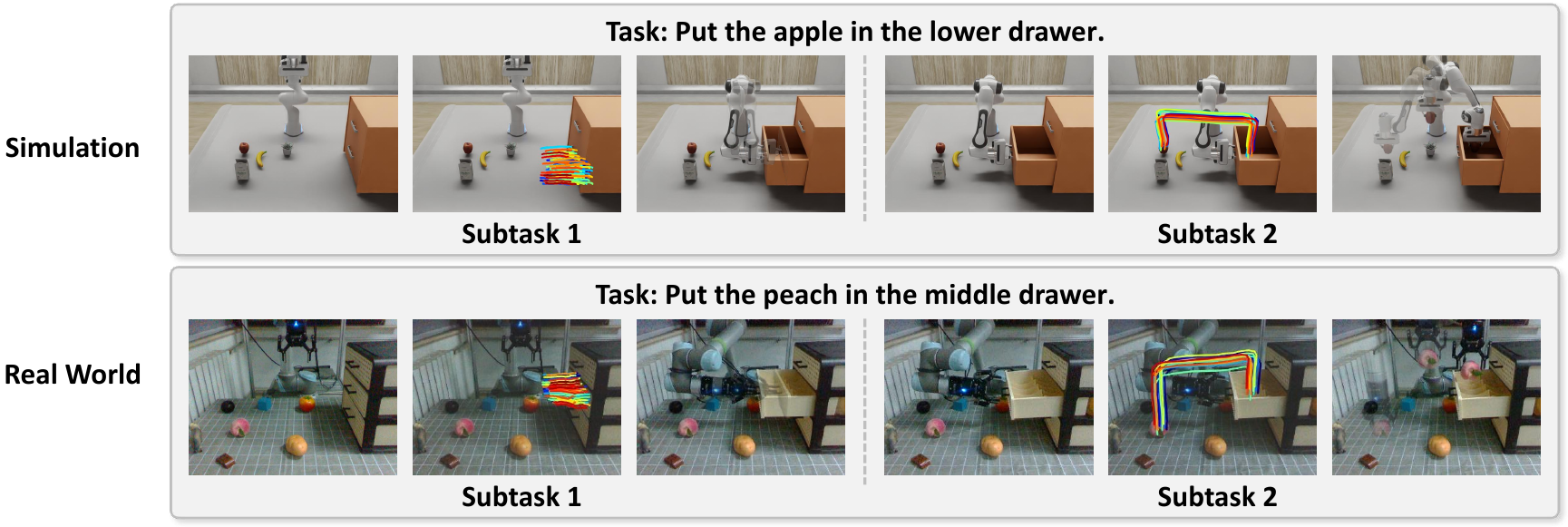}
    \caption{Visualization of the composability of SADiff for long-horizon tasks. We assess the model’s capability to execute multi-stage manipulation sequences by decomposing complex instructions into sequential subtasks.}
    \label{composability_fig}
\end{figure*}

To verify the composability of our approach, we evaluate SADiff on long-horizon tasks that require sequential execution of multiple skills. For a complex task instruction such as \textit{``Put the apple in the lower drawer''}, we employ Qwen-VL~\cite{Qwen-VL} as a VLM-based high-level planner. The planner is provided with the predefined skill set, the current visual observation, and the high-level language instruction, based on which it produces an ordered sequence of subtasks. Subsequently, SADiff generates object-centric motion flow for each subtask and maps it into executable actions before proceeding to the next subtask. As illustrated in the representative rollouts in Fig.~\ref{composability_fig}, this decoupling strategy enables the flexible composition of diverse skills while leveraging the robust generalization capabilities of SADiff to handle task complexity. The successful execution in both simulation and real-world settings demonstrates the seamless composability of our framework and validates its potential for completing complex, long-horizon tasks. 

\section{Conclusion} 
\label{sec:conclusion}

In this work, we propose SADiff, a skill-aware diffusion framework that improves generalization in robotic manipulation by explicitly modeling skill-level information. By integrating a skill-aware encoding module equipped with learnable skill tokens, a skill-constrained diffusion model, and a skill-retrieval transformation strategy, SADiff effectively captures shared skill-level information and systematically integrates it across the encoding, generation, and execution phases. This design enables accurate generation of object-centric motion flow and robust conversion of the predicted 2D flow into executable 3D robot actions across diverse objects and environments. We further construct IsaacSkill, a high-fidelity simulation dataset designed to support skill-centric evaluation and robust sim-to-real transfer. Extensive experiments in both simulation and real-world settings demonstrate that SADiff significantly outperforms existing methods in generalizing to unseen objects, varying environments, and different robot embodiments, while achieving zero-shot sim-to-real transfer without fine-tuning. These results highlight the pivotal role of explicitly modeling skill-level knowledge and integrating it throughout all phases of the pipeline. Moreover, SADiff demonstrates scalability and composability by successfully generalizing to unseen but behaviorally related skills and composing multiple learned skills for long-horizon manipulation.

Although SADiff demonstrates strong generalization, it still faces several limitations. First, the skill taxonomy is manually specified, which limits the expressiveness of the skill space as the framework is extended to more complex or fine-grained manipulation tasks. Future work could explore unsupervised skill discovery, such as using VQ-VAE~\cite{van2017neural} to extract and structure semantic skills from various demonstrations. Second, the 2D motion flow representation struggles to fully resolve spatial dynamics in tasks with out-of-plane rotations (e.g., unscrewing a bottle cap). The proposed skill-aware mechanisms operate on multimodal conditions and skill-level representations rather than being tied to a specific motion representation. Therefore, a natural extension would be to incorporate depth information into the motion representation and generate 3D object motion flow~\cite{zhi20253dflowaction,Huang_2026}.

\bibliographystyle{IEEEtran}
\bibliography{IEEEabrv,references}

\begin{IEEEbiography}[{\includegraphics[width=1in,height=1.25in,clip,keepaspectratio]{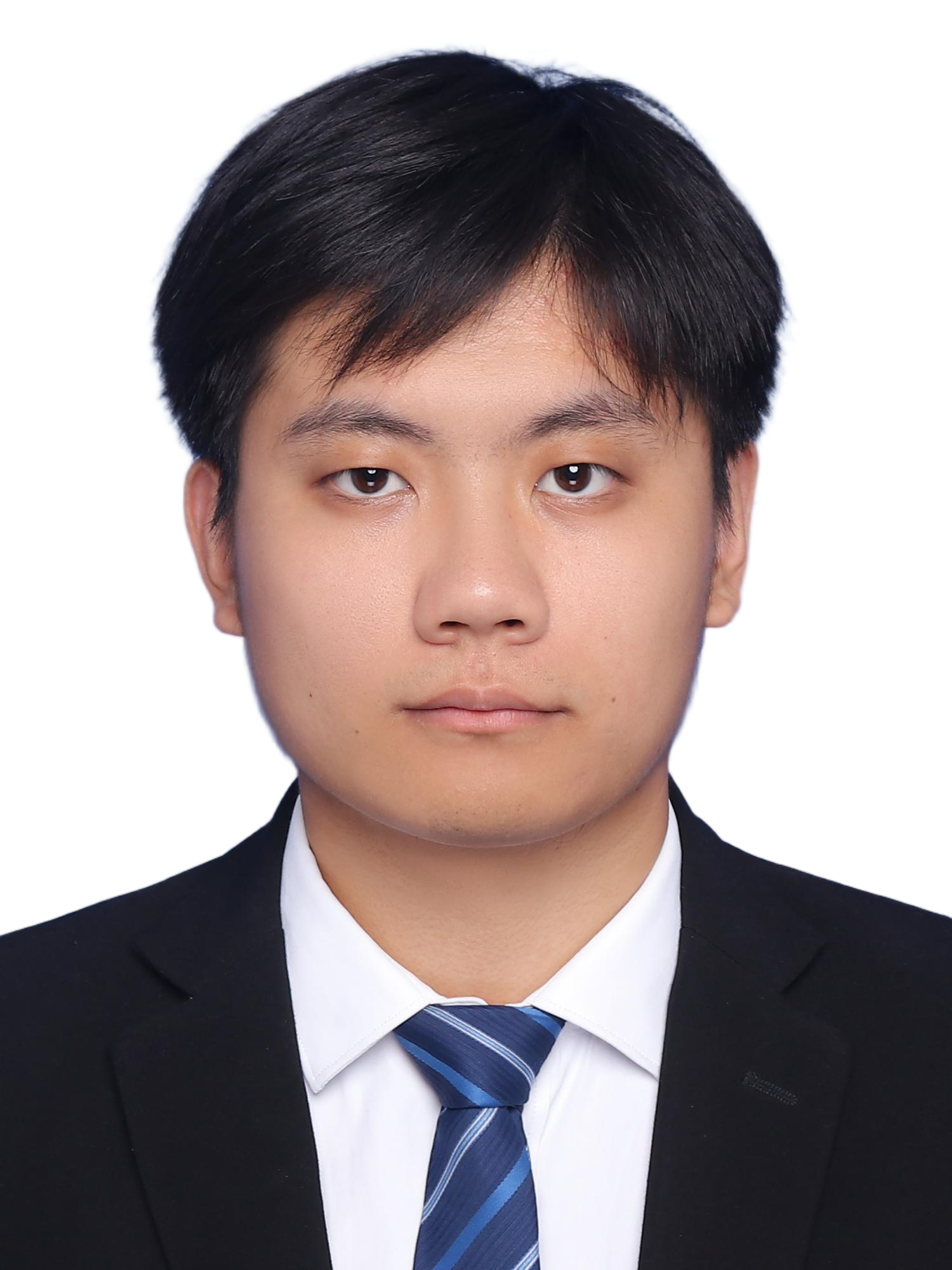}}]{Aoshen Huang}
received the B.Eng. degree in automation from the School of Control Science and Engineering, Shandong University, Jinan, China, in 2024. He is currently pursuing the Ph.D. degree in pattern recognition and intelligent systems at Shandong University. His research interests include robot learning and robotic manipulation.
\end{IEEEbiography}
\begin{IEEEbiography}[{\includegraphics[width=1in,height=1.25in,clip,keepaspectratio]{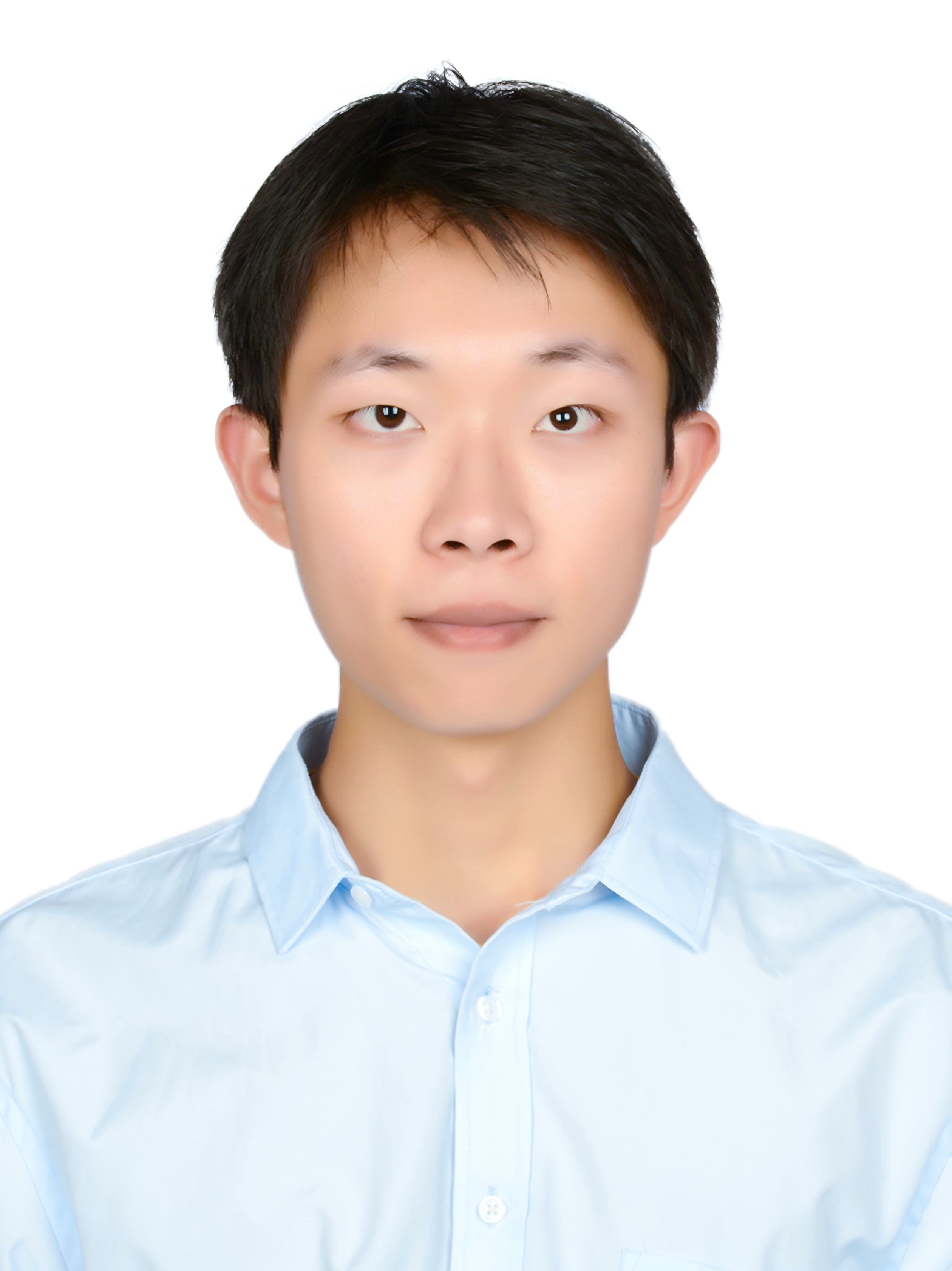}}]{Jiaming Chen}
(Graduate Student Member, IEEE) received the B.Eng. degree in automation and the M.Eng. degree in control science and engineering from Shandong University, Jinan, China, in 2021 and 2024, respectively. He is currently pursuing the Ph.D. degree in computer science at The University of Manchester, U.K. His research interests include computer vision, multimodal learning, and robot learning.
\end{IEEEbiography}
\begin{IEEEbiography}[{\includegraphics[width=1in,height=1.25in,clip,keepaspectratio]{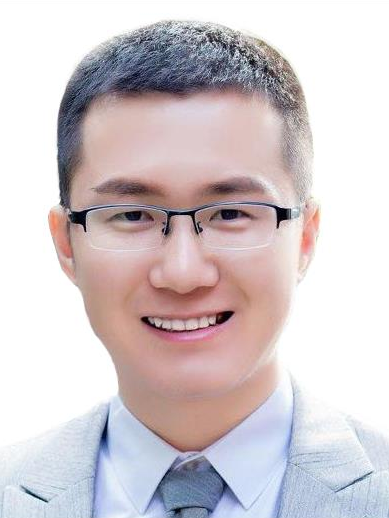}}]{Jiyu Cheng}
(Member, IEEE) received the B.E. degree in automation from Shandong University, Jinan, China, in 2015, and the Ph.D. degree in electronic engineering from The Chinese University of Hong Kong, Hong Kong, in 2019. He is currently an Associate Professor with the School of Control Science and Engineering, Shandong University,  Jinan, China. His research interests include multirobot systems, robot perception, and skill learning.
\end{IEEEbiography}
\begin{IEEEbiography}[{\includegraphics[width=1in,height=1.25in,clip,keepaspectratio]{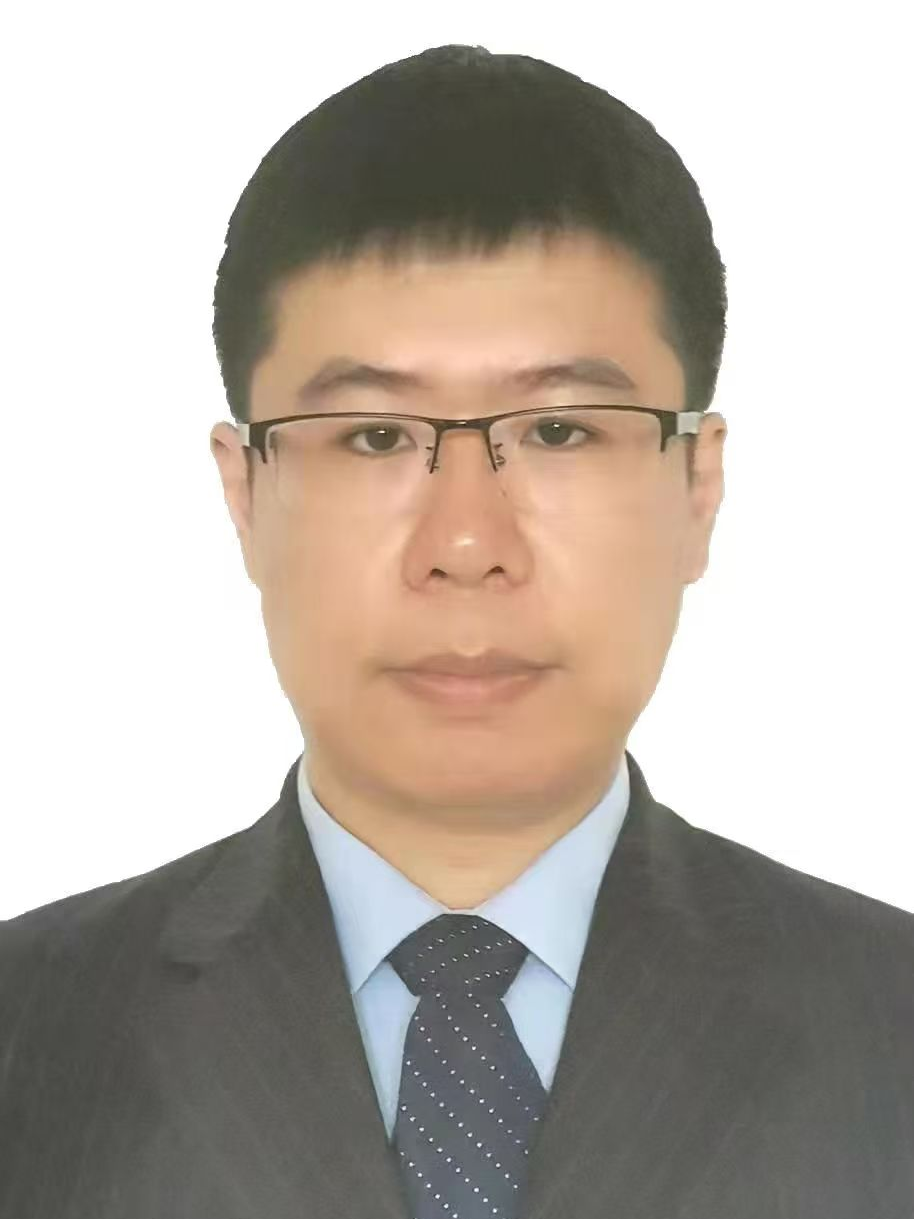}}]{Ran Song}
(Senior Member, IEEE) received the B.Eng. degree in communications engineering from Shandong University, Jinan, China, in 2005, and the Ph.D. degree in electronic engineering from the University of York, York, U.K., in 2009. He is currently a Professor with the School of Control Science and Engineering, Shandong University, Jinan, China. His research interests include robot learning and 3D vision for robotics.
\end{IEEEbiography}
\begin{IEEEbiography}[{\includegraphics[width=1in,height=1.25in,clip,keepaspectratio]{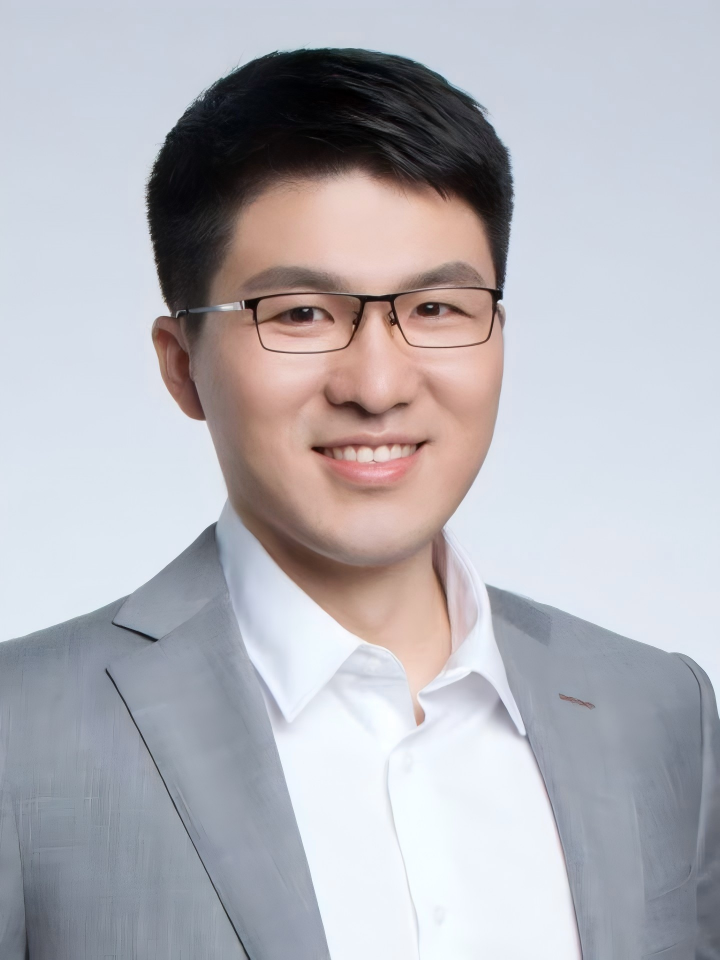}}]{Wei Pan}
(Member, IEEE) received the Ph.D. degree in bioengineering from Imperial College London, London, U.K., in 2016. He is currently a Full Professor of Autonomous Systems and Automatic Control with the School of Engineering, Newcastle University, Newcastle upon Tyne, U.K. He has a broad interest in robot control using Bayesian machine learning and the principles of dynamic control.
\end{IEEEbiography}
\begin{IEEEbiography}[{\includegraphics[width=1in,height=1.25in,clip,keepaspectratio]{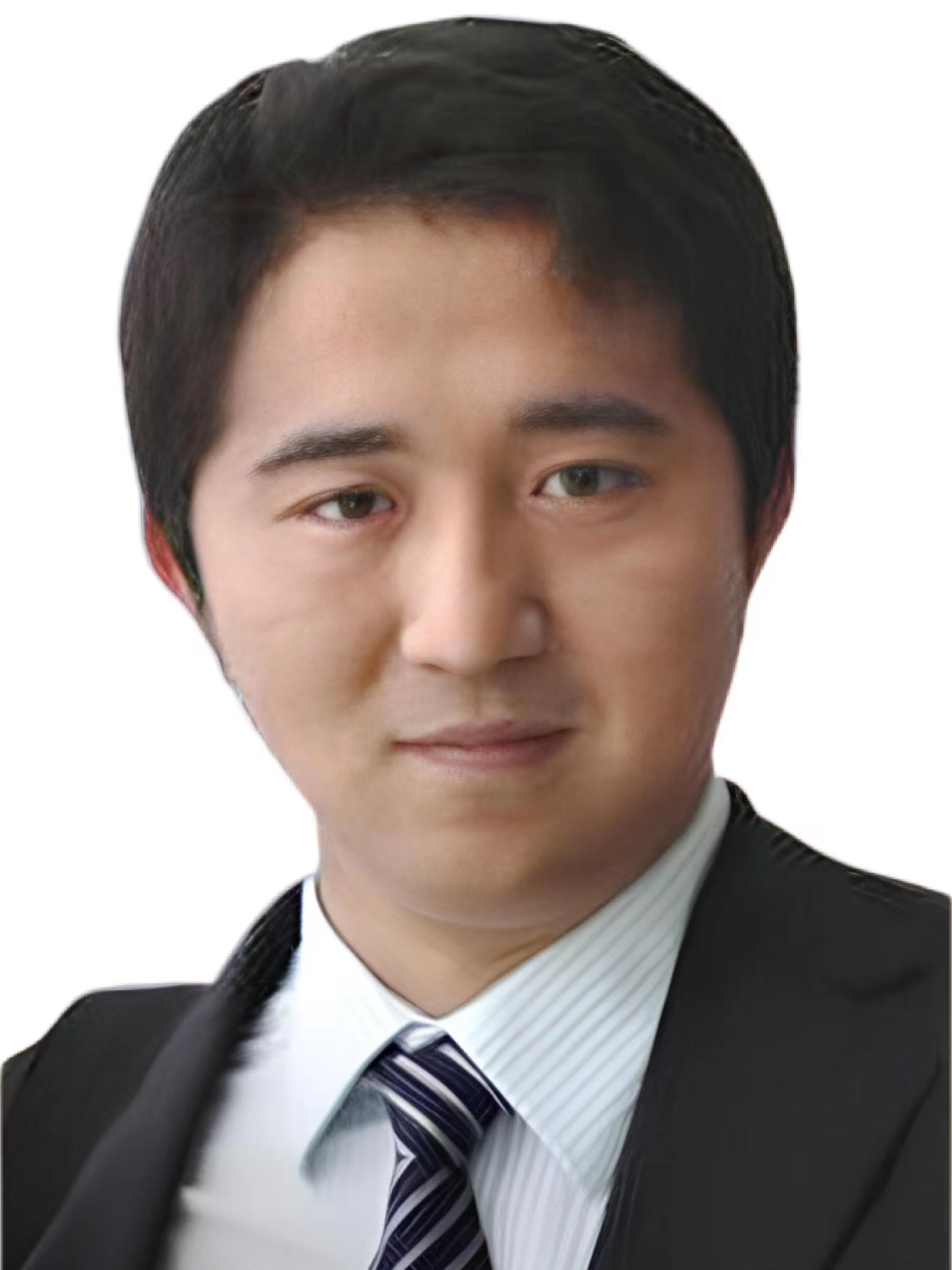}}]{Wei Zhang}
(Senior Member, IEEE) received the Ph.D. degree in electronic engineering from The Chinese University of Hong Kong, Hong Kong, in 2010. He is currently a Professor with the School of Control Science and Engineering, Shandong University, Jinan, China. His research interests include computer vision and robotics. He has served as a program committee member and a reviewer for various international conferences and journals.
\end{IEEEbiography}

\end{document}